\documentclass{article}

\usepackage{microtype}
\usepackage{graphicx}
\usepackage{subfigure}
\usepackage{booktabs} 

\usepackage{hyperref}

\usepackage{qiangstyle}
\usepackage{natbib}
\usepackage{bibentry}

\usepackage[accepted]{icml2018}

\usepackage{times}

\usepackage{algorithm}
\usepackage{algorithmic}
\usepackage{amsmath} 
\usepackage{enumerate}
\usepackage{dsfont}
\usepackage{xcolor}
\usepackage{mathrsfs}

\newcommand{\prop}{\rho}
\renewcommand{\F}{\mathcal{F}}

\begin{document}

\icmltitlerunning{Gradient-Free SVGD}

\twocolumn[
\icmltitle{Stein Variational Gradient Descent Without Gradient}

\begin{icmlauthorlist}
\icmlauthor{Jun Han}{han}
\icmlauthor{Qiang Liu}{liu}
\end{icmlauthorlist}

\icmlaffiliation{han}{Computer Science, Dartmouth College}
\icmlaffiliation{liu}{Computer Science, The University of Texas at Austin}

\icmlcorrespondingauthor{Jun}{jun.han.gr@dartmouth.edu}
\icmlcorrespondingauthor{Qiang}{lqiang@cs.utexas.edu}

\vskip 0.3in
]
\printAffiliationsAndNotice{}

\begin{abstract}
Stein variational gradient decent (SVGD) has been shown to be a powerful approximate inference algorithm for complex distributions. However, the standard SVGD requires calculating the gradient of the target density and cannot be applied when the gradient is unavailable. 
In this work, we develop a gradient-free variant of SVGD (GF-SVGD), which replaces the true gradient with a surrogate gradient, and corrects the induced bias by re-weighting the gradients in a proper form. 
We show that our GF-SVGD can be viewed as the standard SVGD with a special choice of kernel, 
and hence directly inherits the theoretical properties of SVGD.
We shed insights on the empirical choice of the surrogate gradient and propose an annealed GF-SVGD that leverages the idea of simulated annealing to improve the performance on high dimensional complex distributions. 
Empirical studies show that our method consistently outperforms a number of recent advanced gradient-free MCMC methods. 
\end{abstract}

\section{Introduction}
Approximate inference of complex distributions is a long-standing fundamental computational task in machine learning and statistics. 
Traditional methods are based on either Markov chain Monte Carlo (MCMC) \citep{neal2011mcmc, hoffman2014no}, or variational inference (VI) \citep{blei2017variational, zhang2017advances}, both of which have their own inherent pros and cons: MCMC is theoretically sound and asymptotically consistent, but is often slow to converge in  practice; VI is practically faster but has been known to lack theoretical consistency guarantees. 

Stein variational gradient descent (SVGD) \citep{liu2016stein} is recently developed to integrate the advantages of MCMC and VI. 
By leveraging a new type of functional gradient descent of KL divergence on the space of distributions, 
SVGD directly fits a set of non-parametric particles to the distribution of interest, without introducing additional parametric forms. This makes SVGD both theoretically sound \citep{liu2017stein}, practically fast and sample-efficient thanks to its optimization-based formulation. 

Unfortunately, gradient information of the target distribution is not always available in practice. 
In some cases, the distribution of interest is only available as a black-box density function and the gradient cannot be calculated analytically; in other cases, it may be computationally too expensive to calculate the gradient \citep{beaumont2003estimation, andrieu2009pseudo, filippone2014pseudo}. The goal of this paper is to extend SVGD to gradient-free settings. 

\textbf{Our Results}~ 
We propose a simple gradient-free variant of SVGD that replaces the true gradient with a surrogate gradient and uses a form of importance weighting to correct the bias introduced by the surrogate gradient.  
To give a very quick overview of our method, recall that the standard SVGD 
approximates a given distribution $p$ on $\R^d$ with a set of particles $\{\vx_i\}_{i=1}^n$ iteratively updated by 
$\vx_i \gets \vx_i  + \frac{\epsilon}{n} \Delta \vx_i$, where 
\begin{align}\label{update11}
\!\!\!\!\! \Delta \vx_i =\sum_{j=1}^n [\nabla \log p(\vx_j)k(\vx_j, \vx_i) +  \nabla_{\vx_j}k(\vx_j,\vx_i)], 
\end{align}
where $k(\vx,\vx')$ is any positive definite kernel; the term with gradient $\nabla\log p$ drives the particles to the high probability regions of $p$,
and the term with $\nabla k(\vx, \vx')$ acts as a repulsive force to keep the particles away from each other to quantify the uncertainty. 

\newcommand{\scalemath}[2]{\scalebox{#1}{\mbox{\ensuremath{\displaystyle #2}}}}
Our gradient-free SVGD is based on the following generalization of the SVGD update rule: 
\begin{align}
\scalemath{0.99}{
\label{update222}
\!\!\!\!\!\! \Delta \vx_i \propto
\!\sum_{j=1}^n 
\!w_j \big[\nabla \log \rho(\vx_j) k( \vx_j, \vx_i) + \nabla_{\vx_j} k(\vx_j, \vx_i) \big], 
}
\end{align}
which replaces the true gradient $\nabla \log p$ with a surrogate gradient $\nabla\log \rho$ of an arbitrary auxiliary distribution $\rho(\vx)$, and then uses an importance weight $w_j:=\rho(\vx_j)/p(\vx_j)$ to correct the bias introduced by the surrogate.
Perhaps surprisingly, 
we show that the new update can be derived as a standard SVGD update by using an importance weighted kernel $w(\vx)w(\vx')k(\vx,\vx')$, and hence immediately inherits the theoretical proprieties of SVGD; 
for example, particles updated by \eqref{update222} can be viewed as a gradient flow of KL divergence similar to the original SVGD \citep{liu2017stein}. 

The performance of the gradient-free update \eqref{update222} critically depends on the choice of $\rho$. We study this problem empirically and show that it works efficiently, sometimes even better than the original SVGD, if we take $\rho$ to be an \emph{over-dispersed} estimate of $p$ that covers the support of $p$ well. We further improve the algorithm by combining the gradient-free update \eqref{update222} with simulated annealing, which we show consistently outperforms 
a number of other advanced gradient-free Monte Carlo methods, including a gradient-free variant of annealed importance sampling \citep{neal2001annealed}, kernel adaptive Metropolis-Hastings (KAMH) \citep{sejdinovic2014kernel} and kernel  Hamiltonian  Monte Carlo  (KHMC) \citep{strathmann2015gradient}. 

{\bf Discussion and Related Works} 
Almost all gradient-free sampling methods employ some auxiliary (or proposal) distributions that are different, but sufficiently close to the target distribution, followed with some mechanisms to correct the bias introduced by the surrogate distribution. 
There have been a few number of bias-correction mechanisms underlying most of the gradient-free methods, 
including importance sampling, rejection sampling and the Metropolis-Hastings rejection trick. 
The state-of-the-art gradient-free sampling methods are often based on adaptive improvement of the proposals  
when using these tricks, this includes adaptive importance sampling and rejection sampling \citep[]{gilks1992adaptive, cappe2008adaptive, cotter2015parallel, han2017stein}, and adaptive MCMC methods \citep[e.g.,][]{sejdinovic2014kernel, strathmann2015gradient}.  
 
Our method is significantly different from these gradient-free sampling algorithms aforementioned in principle, with a number of key advantages. 
Instead of correcting the bias by either re-weighting or rejecting the samples from the proposal, which unavoidably reduces the effective number of usable particles,  
our method re-weights the SVGD gradient and steers the update direction of the particles in a way that compensates the discrepancy between the target and surrogate distribution, without directly reducing the effective number of usable particles. 

In addition, 
while the traditional importance sampling and rejection sampling methods require the proposals to be simple enough to draw samples from, 
our update does not require to draw samples from the surrogate $\rho$. We can set $\rho$ to be arbitrarily complex as long as we can calculate  $\rho(\vx)$ and its gradient.  
In fact, $\rho(\vx)$ does not even have to be a normalized probability, sidestepping the difficult problem of calculating the normalization constant.

{\bf Outline} The paper is organized as follows. 
Section 2 introduces the background of SVGD. 
Section~3 introduces our novel gradient-free SVGD algorithm, and Section~4 further proposes annealed gradient-free SVGD that combines the advantage of simulated annealing to achieve better performance for complex distributions. 
We present empirical results in Section~5 and conclude the paper in Section~6. 

\section{Stein Variational Gradient Descent}
\label{sec:svgd}
Stein variational gradient descent (SVGD) \citep{liu2016stein} is a 
nonparametric variational inference algorithm that 
iteratively transports a set of particles 
to approximate a given target distribution by performing a type of functional gradient descent on the KL divergence.  
We give a quick overview of its main idea in this section.  

Let $p(\vx)$ be a positive density function on $\R^d$ which we want to approximate with a set of particles 
$\{ \vx_i\}_{i=1}^n$. 
SVGD starts with a set of initial particles $\{  \vx_i\}_{i=1}^n$, 
and updates the particles iteratively by 
\begin{align}\label{equ:xxii}
\vx_i  \gets \vx_i +  \epsilon \ff(\vx_i),  ~~~~ \forall i = 1, \ldots, n,  
\end{align}
where $\epsilon$ is a step size, and 
$\ff\colon \RR^d \to \RR^d$ is a velocity field which should be chosen to drive the particle distribution closer to the target.
Assume the distribution of the particles at the current iteration is $q$, 
and $q_{[\epsilon\ff]}$ is the distribution of the updated particles $\vx^\prime = \vx + \epsilon \ff(\vx)$. 
The optimal choice of $\ff$ can be framed into the following optimization problem:  
\begin{align}\label{equ:ff00}
\ff^* =   \argmax_{\ff \in \F}  \bigg\{  -   \frac{d}{d\epsilon} \KL(q_{[\epsilon\ff]} ~|| ~ p) \big |_{\epsilon = 0}  \bigg\}, 
\end{align}
where $\F$ is the set of candidate velocity fields, and  $\ff$ is chosen in $\F$ to maximize the decreasing rate on the KL divergence between the particle distribution and the target. 

In SVGD, $\F$ is chosen to be the unit ball of a vector-valued reproducing kernel Hilbert space (RKHS) $\H = \H_0 \times \cdots \times \H_0$,
where  $\H_0$ is a RKHS formed by scalar-valued functions associated with a positive definite kernel $k(\vx,\vx')$, that is, 
$\F = \{\ff \in \H \colon ||\ff||_{\H}\leq  1 \}.$
This choice of $\F$ allows us to 
consider velocity fields in infinite dimensional function spaces while still obtaining computationally tractable solution. 

A key step towards solving \eqref{equ:ff00} is to observe that the objective function 
in \eqref{equ:ff00} is a simple linear functional of $\ff$ that connects to Stein operator \citep{oates2017control, gorham2015measuring, liu2016stein, gorham2017measuring, chen2018stein}, 
\begin{align}\label{equ:klstein00}
&~ - \frac{d}{d\epsilon} \KL(q_{[\epsilon\ff]} ~|| ~ p) \big |_{\epsilon = 0}  = \E_{x\sim q}[\steinpxtransp  \ff(\vx)], \\[.5\baselineskip]
&\!\!\!\!\!\!\!\text{with}~~~ \steinpxtransp \ff(\vx)  = \nabla_\vx \log p(\vx) ^\top \ff (\vx)+ \nabla_{\vx}^\top\ff(\vx),  
\end{align}
where $\steinpx$ is a linear operator 
called \emph{Stein operator} and is formally viewed as a column vector similar to the gradient operator $\nabla_{\vx}$. 
The Stein operator $\steinpx$ is connected to Stein's identity which shows that the RHS of \eqref{equ:klstein00} is zero if $p = q$: 
\begin{align}\label{equ:steinid}
\E_{\vx\sim p}[\steinpxtransp \ff(\vx)]  = 0. 
\end{align}
This corresponds to $\frac{d}{d\epsilon} \KL(q_{[\epsilon\ff]} ~|| ~ p) \big |_{\epsilon = 0} = 0$ since there is no way to further decease the KL divergence when $p=q$. 
Eq. \eqref{equ:steinid} is a simple result of integration by parts assuming the value of $p(\vx)\ff(\vx)$ vanishes on the boundary of the integration domain.   

Therefore, the optimization in \eqref{equ:ff00} reduces to 
\begin{align}
\label{solvksd}
\!\!\!\!\D_{\F}(q || p) \overset{def}{=} 
\max_{\ff \in \F} \left\{ \E_{\vx \sim q} \left [\steinpxtransp \ff(\vx)\right] \right\}, 
\end{align}
where $\mathbb{D}_\F(q ~||~ p)$ is the kernelized Stein discrepancy (KSD) defined in \citet{liu2016kernelized,  chwialkowski2016kernel}. 

Observing that \eqref{solvksd} 
is ``simple'' in that it is a linear functional optimization on a 
unit ball of a Hilbert space, \citet{liu2016stein} showed that \eqref{equ:ff00} has a simple closed-form solution: 
\begin{align}\label{svgdoptimal} 
\ff^*(\vx') \propto  \E_{\vx\sim q}[\steinpx k(\vx, \vx')], 
\end{align} 
where $\steinpx$ is applied to variable $\vx$, and 
\begin{equation}
\label{ksddef}
\mathbb{D}^2_\F(q\mid\mid p) = \E_{\vx, \vx'\sim q}[\kappa_p (\bd{x},  \bd{x}')],
\end{equation}
where $\kappa_p (\bd{x},  \bd{x}') := (\stein_p')^\top(\steinpx  k(\vx, \vx'))$ and $\newsteinpx$ denotes the Stein operator applied on variable $\vx'$. 
Here $\kappa_p(\bd x, \bd x')$ 
can be calculated explicitly in Theorem~3.6 of \citet{liu2016kernelized}. 

The Stein variational gradient direction $\ff^*$ provides a theoretically optimal direction that drives the particles towards the target $p$ as fast as possible. In practice, SVGD approximates $q$ using a set of particles, yielding  update rule  \eqref{update11}. 

\section{Gradient-Free SVGD}
The standard SVGD requires the gradient of the target $p$ and cannot be applied when the gradient is unavailable. In this section, we propose a gradient-free variant of SVGD which replaces the true gradient with a surrogate gradient and corrects the bias introduced using an importance weight. 
We start with introducing a  gradient-free variant of Stein's identity and Stein discrepancy.

\paragraph{Gradient-Free Stein's Identity and Stein Discrepancy} 
We can generalize Stein's identity to make it depend on a surrogate gradient $\nabla_\vx \log \prop$ of an arbitrary auxiliary distribution $\rho$, instead of the true gradient $\nabla_\vx \log p$. The idea is to use importance weights to transform Stein's identity of $\prop$ into an identity regarding $p$: 
recall the Stein's identity of $\prop$:  
$$\E_{\vx\sim \prop} [ \steinbxtransp \ff(\vx)  ] =  0.$$ 
It can be easily seen that it is equivalent to the following \emph{importance weighted Stein's identity}: 
\begin{align} \label{equ:iwstein}
\E_{\vx\sim p} \bigg[\frac{\prop(\vx)}{p(\vx)} \steinbxtransp \ff(\vx) \bigg] = 0, 
\end{align}
which is already gradient free since it depends on $p$ only through the value of $p(\vx)$, not the gradient. 
\eqref{equ:iwstein} holds for an arbitrary auxiliary distribution $\rho$ which satisfies $\rho(\vx)/p(\vx) <\infty$ for any $\vx$.

Based on identity \eqref{equ:iwstein}, it is straightforward to define an 
importance weighted Stein discrepancy 
\begin{align}\label{equ:grds}
\D_{\F,\prop}(q~||~p) = \max_{\ff \in \F}\bigg\{ \E_{x\sim q}\bigg[\frac{\prop(\vx)}{p(\vx)} \steinbxtransp \ff(\vx) \bigg] \bigg\}, 
\end{align}
 which is gradient-free if $\rho$ does not depend on the gradient of $p$. 
Obviously, this includes the standard Stein discrepancy in Section~\ref{sec:svgd} as special cases: 
if $\prop = p$, then $\D_{\F,\rho}(q~||~p) = \D_\F(q~||~p)$, reducing to the original definition in \eqref{solvksd}, while if $\prop = q$, then $\D_{\F, \rho}(q~||~p) = \D_{\F}(p~||~q)$, which switches the order of $p$ and $q$.

It may appear that $\D_{\F, \rho} (q~||~p)$ strictly generalizes the definition~\eqref{solvksd} of Stein discrepancy. 
One of our key observations, however, shows that this is not the case. Instead, 
$\D_{\F, \rho} (q~||~p)$ can also be viewed as a special case of 
$\D_{\F} (q~||~p)$, 
by replacing $\F$ in \eqref{solvksd} with 
$$w \F := \{w(\vx) \ff(\vx) \colon \ff\in \F\},$$
where $w(\vx) = {\prop(\vx)}/{p(\vx)}$. 
\begin{thm}\label{pro:wphi}
Let $p$, $\prop$ be  positive differentiable densities and $w(\vx) = {\prop(\vx)}/{p(\vx)}$. We have 
\begin{align}\label{equ:ws}
 w(\vx)\steinbxtransp\ff(\vx)  = \steinpxtransp \big(w(\vx)\ff (\vx) \big).
\end{align}
Therefore,
$\S_{\F, \prop}(q~||~p)$ in \eqref{equ:grds}
is equivalent to 
\begin{align}
 \S_{\F, \prop}(q~||~p)  
 &  = \max_{\ff \in \F}\big\{ \E_{\vx\sim q} [\steinpxtransp \big( w(\vx)\ff(\vx)\big)]\big\} \label{sbf} \\
 & = \max_{\ff \in w\F} \big \{\E_{\vx\sim q}[\steinpxtransp \ff(\vx)]     \big \}  \label{sbf2}\\
 & = \S_{w\F}(q~||~p). \notag
\end{align}
\end{thm}
{\bf Proof:} The proof can be found in the appendix A. \hfill $\square$

Identity~\eqref{equ:ws} is interesting because it is \emph{gradient-free} (in terms of $\nabla_\vx\log p$) from the left hand side,
but \emph{gradient-dependent} from the right hand side; 
this is because the $\nabla_\vx \log p$ term in $\steinpx$ is cancelled out when applying $\steinpx$ on the density ratio $w(\vx) = \rho(\vx)/p(\vx)$.

It is possible to possible to further extend our method to take $\rho(\vx)$ and 
$w(\vx)$ to be general \emph{matrix-valued} functions,  
in which case the operator $\steinpxtransp (w(\vx)\phi(\vx))$ 
is called diffusion Stein operator in 
\citet{gorham2016measuring}, corresponding to various forms of Langevin diffusion when taking special values of $w(\vx)$.    
We leave it as future work to explore 
 $\rho$. 

\paragraph{Gradient-Free SVGD}
Theorem~\ref{pro:wphi} suggests that by simply multiplying $\ff$ with an importance weight $w(\vx)$ (or replacing $\F$ with $w\F$), one can transform Stein operator $\steinpx$ to operator $\steinbx$, which depends on $\nabla \log \rho$ instead of $\nabla \log p$  (\emph{gradient-free}). 
\begin{algorithm}[t] %
\caption{Gradient-Free SVGD (GF-SVGD)}  
\label{alg:alg1}
\begin{algorithmic}
\STATE {\bf Input:} 
Target distribution $p(\bd{x})$; 
Surrogate $\prop(\bd{x})$ and its score function $\bd{s}_{\prop}(\vx) := \nabla_\vx\log \prop.$ 
\STATE {\bf Goal:} Find particles  $\{\bd{x}_i\}_{i=1}^n$ to approximate $p.$ 
\STATE {\bf Initialize} particles $\{\vx^0_i\}_{i=1}^n$ from any distribution $q$.
\FOR{iteration $\ellt$}
\vspace{-1.\baselineskip}
\STATE
\hspace{-0.4\baselineskip}
\begin{align}
& \bd{x}_i^{\ellt+1} ~ \leftarrow ~ \bd{x}_i^\ellt  ~  + ~ \Delta\vx_i^\ellt, 
~~~\forall i = 1, \ldots, n,  ~~\text{where}~ \notag \\ 
&
\!\!\!\!\!\!\!\!\!\!\!\Delta\vx_i^\ellt = 
\frac{\epsilon_{\ellt,i}}{
Z_\ellt}\sum_{j=1}^n
w(\vx_j^\ellt)\big[ \bd{s}_{\prop}(\vx_j^\ellt) k( \vx_j^\ellt,  \vx_i^\ellt) + \nabla_{\vx_j} k(\vx_j^\ellt, \vx_i^\ellt) \big], \notag
\end{align}
where $w(\vx) := \prop(\vx)/p(\vx)$,  $Z_\ellt = \sum_{j=1}^n w(\vx_j^\ellt)$, 
and $\epsilon_{t,i}$ is a step size. 
\ENDFOR
\vspace{-.2\baselineskip}
\end{algorithmic}
\end{algorithm}
This idea can be directly applied to derive a gradient-free extension of SVGD, 
by updating the particles using velocity fields of form $w(\vx)\ff(\vx)$ from  space $w\F$:   
\begin{equation}
\label{gf:closed}
 \vx\gets \vx + \epsilon w(\vx) \ff^*(\vx),
\end{equation}
where $\ff^*$ maximzies the decrease rate of KL divergence, 
\begin{align}
\label{gradfreeKLmin}
\!\!\!\! \ff^*& 
\!=\! \argmax_{\ff \in \H} \!\!\left\{ \E_{q} [\steinpxtransp (w(\vx)\ff(\vx))], \mathrm{s.t.}~||\ff ||_{\H} \leq 1 \!\right\}\!. 
\end{align}
Similar to \eqref{solvksd}, 
we can derive a closed-form solution for \eqref{gradfreeKLmin} when $\H$ is RKHS. 
To do this, it is sufficient to recall that if $\H$ is an RKHS with kernel $k(\vx,\vx')$, then $w\H$ is also an 
RKHS, with an ``importance weighted kernel'' \citep{berlinet2011reproducing} 
\begin{align}\label{tildk}
    \tilde k(\vx,\vx') = w(\vx)w(\vx')k(\vx,\vx').
\end{align}
\begin{thm}
When $\Hd$ is an RKHS with kernel $k(\vx,\vx')$, the optimal solution of \eqref{sbf} is ${\ff}^*/||{\ff}^*||_\Hd,$ where 
\begin{align}
{\ff}^*(\cdot) 
& =\E_{\vx\sim q}[\steinpx(w(\vx)k(\vx, \cdot))] \label{newvel}
\\
& =\E_{\vx\sim q}[w(\vx) \steinbx k(\vx, \cdot)], \label{newvel2}
\end{align}
where the Stein operator $\steinbx$ is applied to variable $\vx$, $\steinbx k(\vx, \cdot)=\nabla_\vx \log \rho(\vx)k(\vx, \cdot)+\nabla_{\vx}k(\vx, \cdot).$
 Correspondingly, the optimal decrease rate of KL divergence in \eqref{gradfreeKLmin} equals the square of $\S_{\F, \rho}(q~||~p)$, which equals 
\begin{equation}
\label{newksd}
 \S_{\F, \prop}(q ~||~ p) = (\E_{\vx, \vx'\sim q}[w(\vx)w(\vx')\kappa_\prop(\vx,\vx')])^{\frac12},
\end{equation}
where $\kappa_\prop(\vx,\vx') = (\newsteinbx)^\top (\steinbx k(\vx,\vx'))$ and $\newsteinbx$ is the Stein operator applied on variable $\vx'$.
\label{theom}
\end{thm}
{\bf Proof:} The proof can be found in the appendix B. \hfill $\square$

The form in \eqref{newksd} allows us to estimate  $\S_{\F, \prop}(q~||~p)$ empirically either using U-statistics or V-statistics when $q$ is observed through an i.i.d. sample, with the advantage of being gradient-free.  
Therefore, it can be directly applied to construct gradient-free methods for goodness-of-fit tests~\citep{liu2016kernelized, chwialkowski2016kernel} and black-box importance sampling~\citep{liu2016black} when the gradient of $p$ is unavailable. 
We leave this to future works. 

Using the gradient-free form of $\ff^*$ in \eqref{newvel2}, 
we can readily derive a gradient-free SVGD update $\vx_i \gets \vx_i  + \Delta \vx_i$, with  
\begin{align}\label{dx}
\Delta \vx_i = 
\frac{\epsilon_i}{Z} \sum_{j=1}^n w(\vx_j) \steinbx k(\vx_j, \vx_i),
\end{align}
where the operator $\steinbx$ is applied on variable $\vx_j$, and we set $Z = n$, viewed as a normalization constant, and $\epsilon_i = \epsilon w(\vx_i)$, viewed as the step size of particle $\vx_i$ .  

Since $\epsilon_i/Z$ is a scalar, we can change it in practice without altering the set of fixed points of the update. 
In practice, because the variability of the importance weight $ w(\vx_i)$ can be very large, making the updating speed of different particles significantly different, we find it is empirically better to determine $\epsilon_i$ directly using off-the-shelf step size schemes such as Adam~\citep{kingma2014adam}. 

In practice, we also replace $Z={n}$ with a self-normalization factor $Z=\sum_{j=1}^n w(\vx_j)$ (see Algorithm~\ref{alg:alg1}). 
We find this makes tuning step sizes become easier in practice, and more importantly, 
avoids to calculate the normalization constant of either $p(\vx)$ or $\rho(\vx)$. 
This sidesteps the critically challenging problem of calculating the normalization constant and allows us to essentially choose $\rho(\vx)$ to be an arbitrary positive differentiable function once we can calculate its value and gradient. 

\paragraph{Choice of the Auxiliary Distribution $\prop(\vx)$}
Obviously, the performance of gradient-free SVGD (GF-SVGD) 
critically depends on the choice of the auxiliary distribution $\prop(\vx)$. 
Theoretically, gradient-free SVGD is just a standard SVGD with the importance weighted kernel $\tilde k(\vx, \vx')$. 
Therefore, the optimal choice of $\prop$ is essentially the problem of choosing an optimal kernel for SVGD, which, unfortunately, is a difficult, unsolved problem. 

In this work, we take a simple heuristic that sets $\prop(\vx)$ to approximate $p(\vx)$. This is based on the justification that if the original $k(\vx,\vx')$ has been chosen to be optimal or ``reasonably well'', 
we should take $\prop(\vx)\approx p(\vx)$ so that $\tilde k(\vx, \vx')$ is close to $k(\vx,\vx')$ and GF-SVGD will have similar performance as the original SVGD. 
In this way, the problem of choosing the optimal auxiliary distribution $\prop(\vx)$ and the optimal kernel $k(\vx, \vx')$ is separated, and different kernel selection methods can be directly plugged into the algorithm.  
In practice, we find that $\prop(\vx)\approx p(\vx)$ serves a  reasonable heuristic when using Gaussian RBF kernel $k(\vx,\vx')$. 
Interestingly, our empirical observation shows that a widely spread $\prop(\vx)$ tends to give better and more stable results than 
peaky $\prop(\vx)$. In particular, Figure~\ref{fig:gfgauss} in the experiment section shows that in the case when both $p(\vx)$ and $\rho(\vx)$ are Gaussian and RBF kernel is used,
the best performance is achieved when the variance of $\prop(\vx)$ is larger than the variance of $p$. 
In fact, the gradient-free SVGD update \eqref{dx}
still makes sense even 
when $\prop(\vx) = 1$, corresponding to an improper distribution with infinite variance: 
\begin{align}\label{invp}
\Delta \vx_i  = \frac{\epsilon_i}{Z} \sum_{j=1}^n \frac{1}{p(\vx_j)} \nabla_{\vx_j} k(\vx_j,\vx_i). 
\end{align}
This update is interestingly simple; it has only a repulsive force and relies on an inverse probability $1/p(\vx)$ to adjust the particles towards the target $p(\vx)$. 
We should observe that it is \emph{as general as} 
the GF-SVGD update \eqref{dx} (and hence the standard SVGD update \eqref{update11}), because if we replace $k(\vx,\vx')$ with $\prop(\vx)\prop(\vx')k(\vx,\vx')$, \eqref{invp} reduces back to \eqref{dx}.    
%
All it matters is the choice of the kernel function. 
With a ``typical'' kernel such as RBF kernel, we empirically find that the particles by the update \eqref{invp} can  estimate the mean parameter reasonably  well (although not optimally), but tend to  overestimate the variance because the repulsive force dominates; see  Figure~\ref{fig:gfgauss}.   

\begin{algorithm}[t]
\caption{Annealed SVGD (A-SVGD)}
\begin{algorithmic}
\STATE {\bf Inputs:} $p(\vx)$, distribution path $\{p_\ellt\}_{\ellt=1}^T$ with $p_T = p.$ 
\STATE {\bf Initialize} particles $\{\vx^0_i\}_{i=1}^n$ from any distribution. 
\FOR{iteration $\ellt =0, \cdots, T-1$}
\STATE Update the particles to get $\{\bd{x}_i^{t+1}\}_{i=1}^n$ by running the typical SVGD with $p_{\ellt+1}$ as the target for $m$ steps. 
\ENDFOR
\STATE {\bf Output:} $\{\bd{x}_i^{T}\}_{i=1}^n$ as an approximation of $p.$ 
\STATE {\bf Remark:} $m=1$ is sufficient when $T$ is large. 
\vspace{-.2\baselineskip}
\end{algorithmic}
\label{alg2}
\end{algorithm}

\section{Annealed Gradient-Free SVGD}
In practice, it may be difficult to directly find $\rho(\vx)$ that closely approximates the target $p$,  
causing the importance weights to have undesirably large variance and deteriorate the performance.  
In this section, we introduce an annealed GF-SVGD algorithm that overcomes the difficulty of choosing $\rho$ and improves the performance 
by iteratively approximating a sequence of distributions which interpolate the target distribution with a simple initial distribution. 
In the sequel, we first introduce the annealed version of the basic SVGD and then its combination with GF-SVGD.

\newcommand{\pz}{p_0}
\textbf{Annealed SVGD (A-SVGD)} is a simple combination of SVGD and simulated annealing, and has been discussed by \citet{liu2017steinpolicy} in the setting of reinforcement learning. 
Let $\pz(\vx)$ be a simple initial distribution. 
We define a path of distributions that interpolate between $\pz(\vx)$ and $p(\vx)$: 
 $$
 p_\ellt(\bd{x}) \propto \pz(\bd{x})^{1-\alpha_\ellt} p(\bd{x})^{\alpha_\ellt},
 $$
where $0= \alpha_0< \alpha_1< \cdots <\alpha_T= 1$ is a set of temperatures. 
Annealed SVGD starts from a set of particle $\{x_i^0\}_{i=1}^n$ drawn from $p_0$, and at the $\ellt$-th iteration, updates the particles so that $\{x_i^{\ellt+1}\}_{i=1}^n$ approximates the intermediate distribution $p_{\ellt+1}$ by running $m$ steps of SVGD with $p_{\ellt+1}$ as the target. See Algorithm~\ref{alg2}. In practice, $m=1$ is sufficient when $T$ is large.
\begin{algorithm}[t] %
\caption{Annealed GF-SVGD (AGF-SVGD)}  
\begin{algorithmic}
\STATE {\bf Input:}  Target distribution $p(\bd{x})$; initial distribution $\pz(\bd{x})$;  intermediate distributions  $\{p_t\}_{t=1}^T$. 
\STATE {\bf Goal:} Particles $\{\bd{x}_i\}_{i=1}^n$ to approximate $p(\bd{x}).$ 
\STATE {\bf Initialize} particles $\{\vx^0_i\}_{i=1}^n$ drawn from $p_0$.  
\FOR{iteration $\ellt =0,\cdots, T-1$}
\vspace{-1.\baselineskip}
\STATE
\hspace{-10\baselineskip}
\begin{align}
& \bd{x}_i^{\ellt+1}  \leftarrow  \bd{x}_i^{\ellt} + \Delta\vx_i^\ellt,  
~~\forall i = 1, \ldots, n,  ~~\text{where}~ \notag \\
& \!\!\!\!\!\!\!\!\!\!\!\!\!\!\!\!\!\!\! \Delta \vx_i^\ellt = 
\frac{\epsilon_{\ellt,i}}{
Z_\ellt}\sum_{j=1}^n
w_j^\ellt \big[ \vv s^\rho_{j, \ellt+1} k( \vx_j^\ellt, \vx_i^\ellt) + \nabla_{\vx_j} k(\vx_j^\ellt, \vx_i^\ellt)  \big], \notag\vspace{-1em} 
\end{align}
where $\vv s^\rho_{j, \ellt+1} = \nabla_\vx \log \rho_{\ellt+1}(\vx_j^\ellt)$ and $\rho_{\ellt+1}$ is defined in \eqref{bt}; $w_j^\ellt = \rho_{\ellt+1}(\vx_j^\ellt)/p_{\ellt+1}(\vx_j^\ellt)$, 
$Z_\ellt = \sum_{j=1}^n w_j^\ellt$. 
\ENDFOR
\STATE {\bf Output:} $\{\bd{x}_i^{T}\}_{i=1}^n$ to approximate $p.$
\vspace{.2\baselineskip}
\end{algorithmic}
\label{alg3}
\end{algorithm}
It is useful to consider the special case when $p_0=const$, 
and hence $p_\ellt(\bd{x}) \propto p(\bd{x})^{\alpha_\ellt}$, yielding an annealed SVGD update of form
$$
\Delta \vx_i
= \frac{\epsilon}{n}\sum_{j=1}^n [\nabla_\vx \log p(\vx_j) k(\vx_{j}, \vx_i) + \frac{1}{\alpha_\ellt}\nabla_{\vx_j} k(\vx_{j}, \vx_i)], 
$$
where the repulsive force is weighted by the inverse temperature $1/\alpha_\ellt$. 
As $\alpha_\ellt$ increases from 0 to 1, the algorithm starts with a large repulsive force and gradually decreases it to match the temperature of the distribution of interest. 
This procedure is similar to the typical simulated annealing, but enforces the diversity of the particles using the deterministic repulsive force, instead of random noise. 

\begin{figure*}[ht]
\centering
\begin{tabular}{ccccc}
\includegraphics[height=0.19\textwidth]{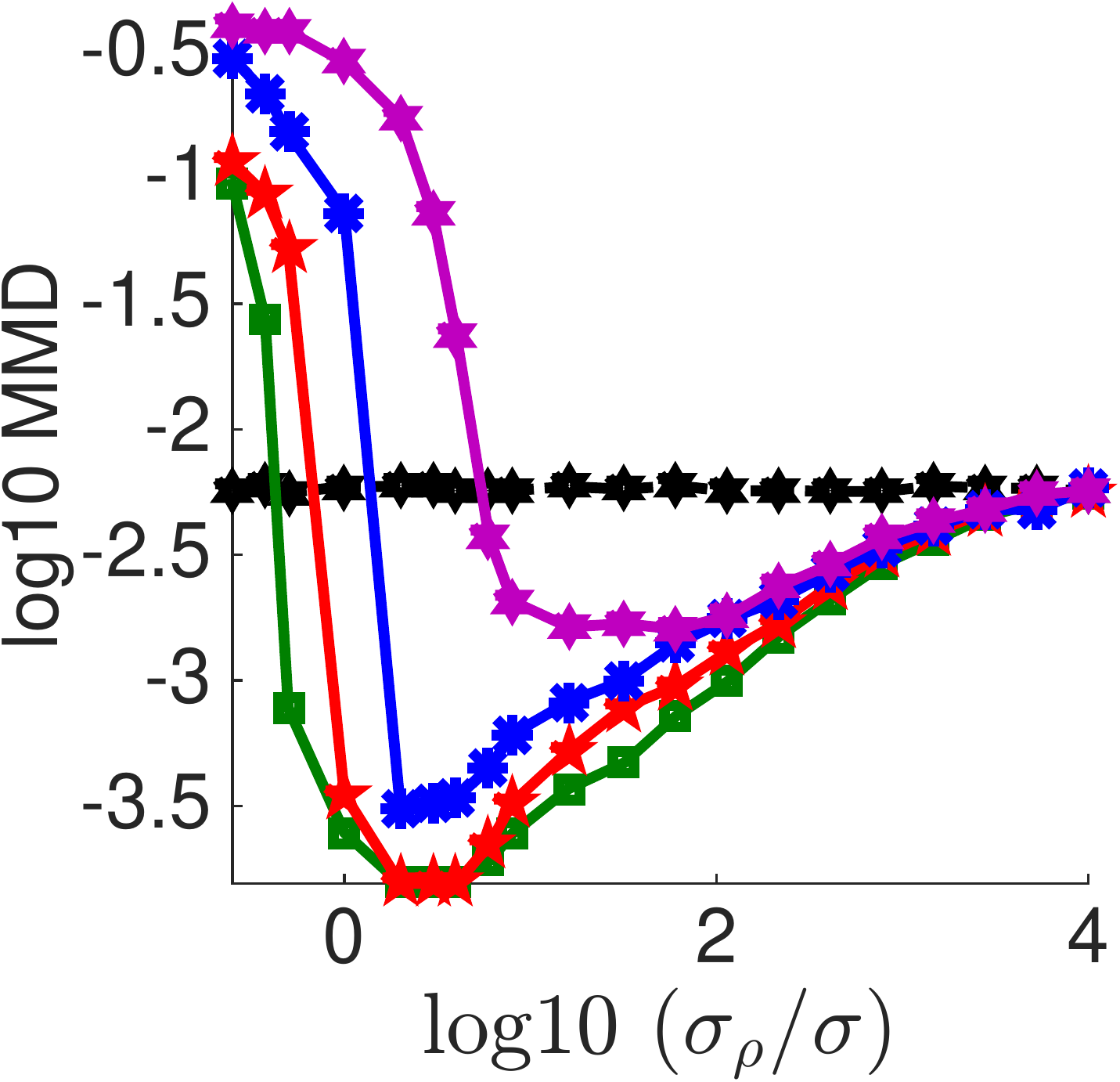}&
\includegraphics[height=0.19\textwidth]{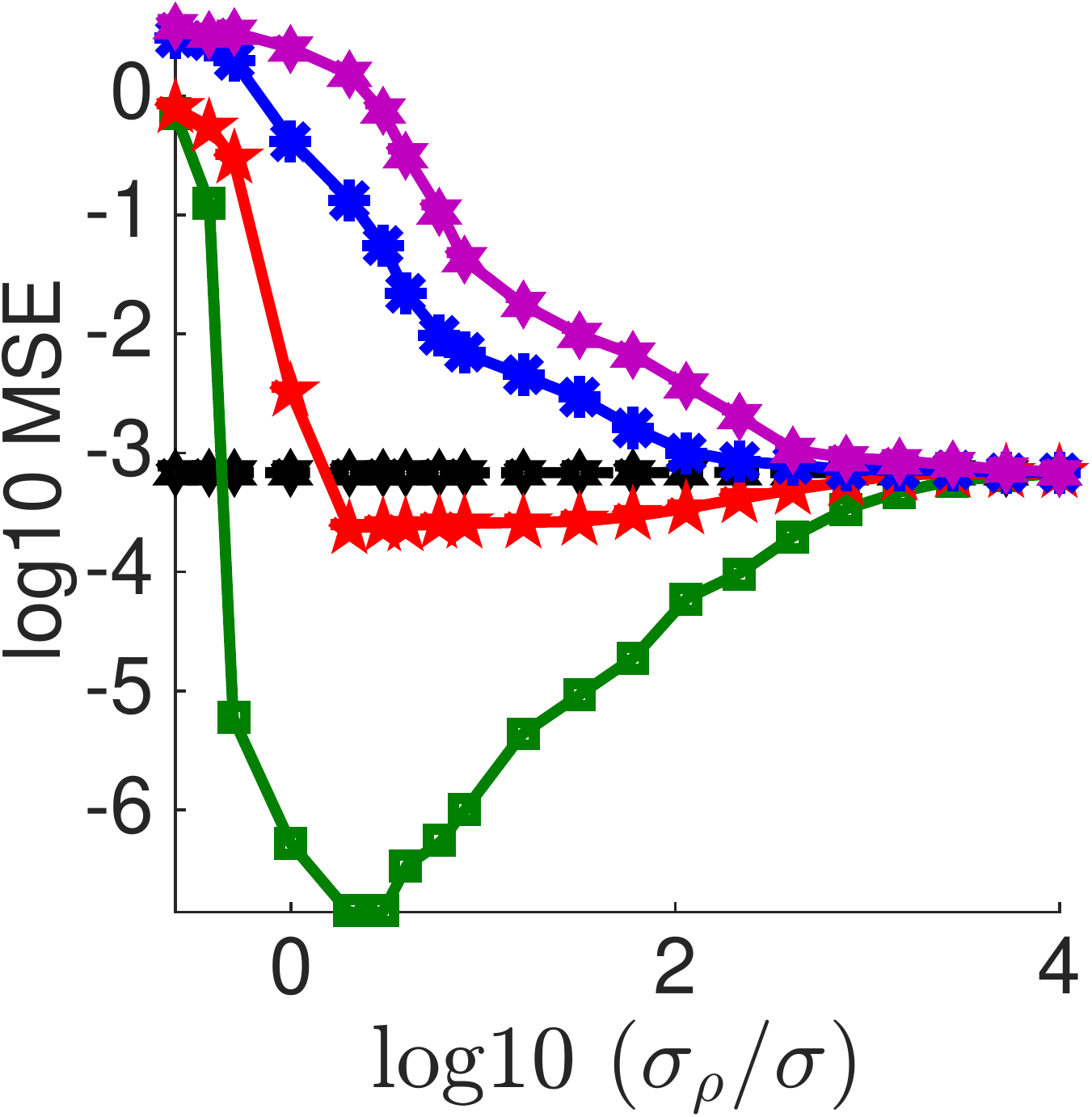} &
\includegraphics[height=0.19\textwidth]{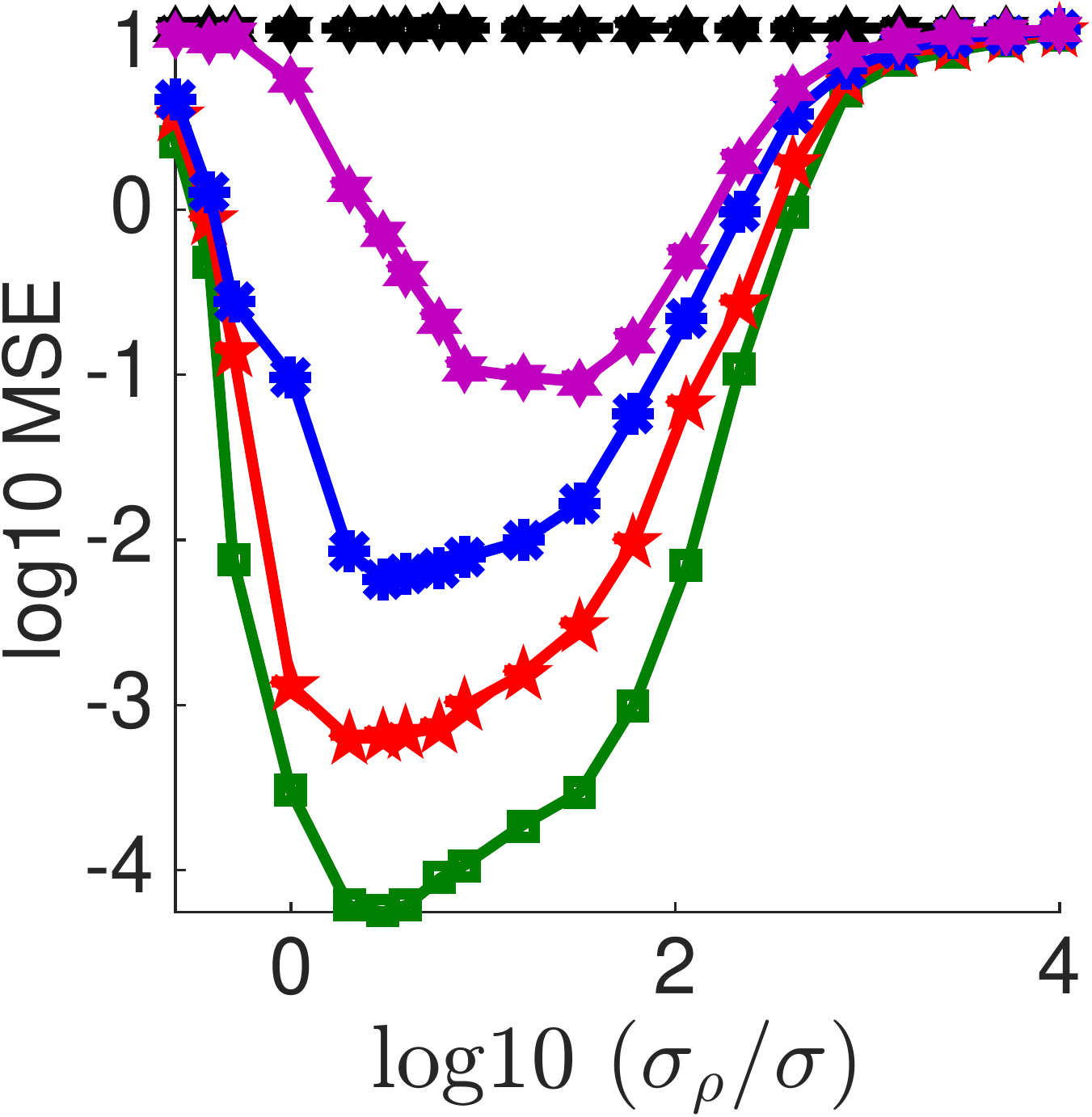} &
\raisebox{2em}{\includegraphics[height=0.1\textwidth]{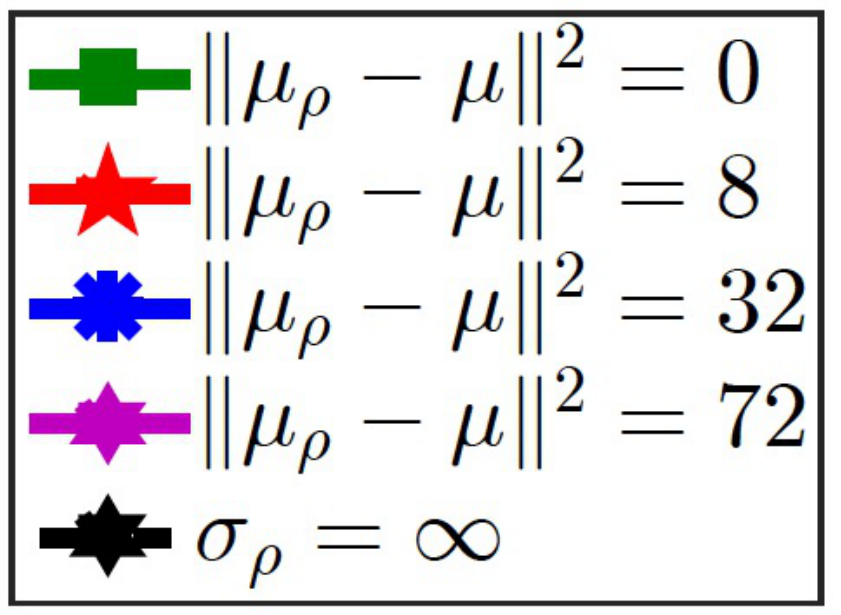}}&
\includegraphics[height=0.19\textwidth]{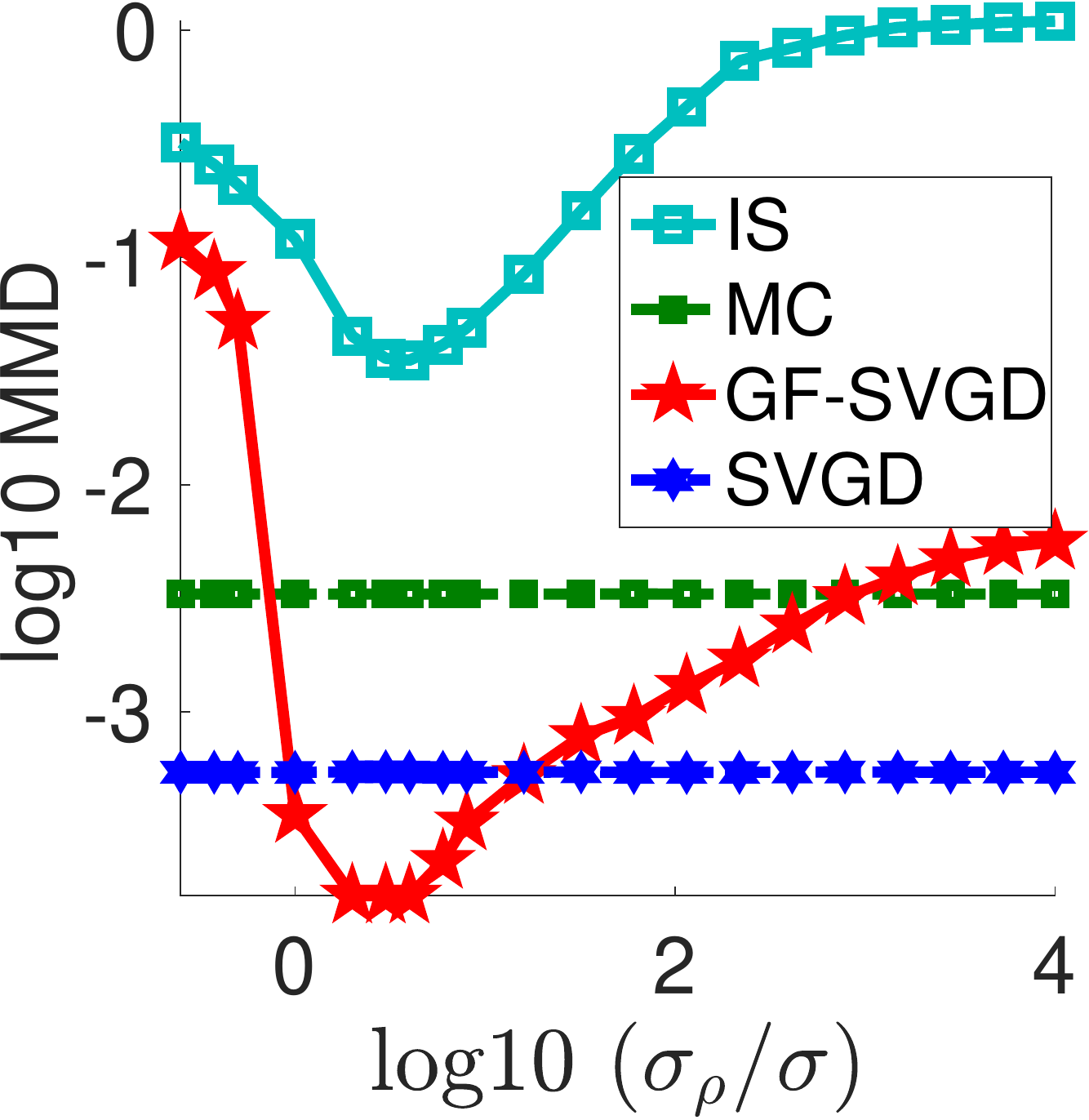} \\
 {\small (a) MMD} & {\small (b) Mean} &  {\small (c) Variance}&~~~ & {\small (d) MMD} \\
\end{tabular}
\caption{\small 
Results of GF-SVGD on 2D multivariate Gaussian distribution as we change the mean $\vv\mu_\rho$ and variance $\sigma_\rho$ of the surrogate $\rho(\vx)$. 
We can see that the best performance is achieved by matching the mean of  $\rho$ and the target $p$  ($\bd{\mu}_\rho = \bd{\mu}$),
and making $\sigma_\rho$ slightly larger than the variance $\sigma$ of $p$ (e.g., 
$\log10(\sigma_\rho/\sigma)\approx 0.5$ or 
$\sigma_\rho \approx 3 \sigma$). 
(d) uses the same setting with $||\vv\mu_\rho-\bd{\mu}||^2=8$, but also adds the result of exact Monte Carlo sampling, gradient-based SVGD, and importance sampling (IS) whose proposal is $\rho$, the same as the auxiliary distribution used by GF-SVGD shown in the red curve. We use $n=100$ particles in  this plot.
}
\label{fig:gfgauss}
\end{figure*}

\newcommand{\kb}{k_\prop}
\textbf{Annealed Gradient-Free SVGD} (AGF-SVGD) is the gradient-free version of annealed SVGD 
which replaces the SVGD update with an GF-SVGD update. 
Specifically, at the $\ellt$-th iteration when we want to update the particles to match $p_{\ellt+1}$, 
we use a GF-SVGD update with auxiliary distribution $\prop_{\ellt+1} \approx p_{\ellt+1}$, which we construct by using a simple kernel curve estimation 
\begin{align}\label{bt}
\prop_{\ellt+1}(\vx) \propto \sum_{j=1}^n p_{\ellt+1}(\vx_j^{\ellt}) \kb(\vx_j^{\ellt}, \vx), 
\end{align}
where $\kb$ is a smoothing kernel (which does not have to be positive definite).  
Although there are other ways to approximate $p_{\ellt+1}$, 
this simple heuristic is computationally fast, and the usage of smoothing kernel makes $\prop_{\ellt+1}$ an \emph{over-dispersed} estimate which we show perform well in practice (see Figure~\ref{fig:gfgauss}). 
Note that here $\rho_{\ellt+1}$ is constructed to  \emph{fit smooth curve $p_{\ellt+1}$}, which leverages the function values of $p_{\ellt+1}(\vx)$ and is insensitive to the actual distribution of the current particles $\{\vx_j^{\ellt}\}$. 
It would be less robust to construct $\rho_{t+1}$ \emph{as a density estimator of distribution $p_{\ellt+1}$} because the actual distribution of the particles may deviate from what we  expect in practice. 

The procedure is organized in Algorithm~\ref{alg3}. Combining the idea of simulated annealing with gradient-free SVGD makes it easier to construct an initial surrogate distribution and estimate a good auxiliary distribution at each iteration, decreasing the variance of the importance weights. We find that it significantly improves the performance over the basic GF-SVGD for complex target distributions.

\section{Empirical Results} 
We test our proposed algorithms on both synthetic and real-world examples. 
We start with testing our methods on simple multivariate Gaussian and Gaussian mixture models, developing insights on the optimal choice of the auxiliary distribution. 
%
We then test AGF-SVGD on Gaussian-Bernoulli restricted Boltzmann machine (RBM)
and compare it with advanced gradient-free MCMC such as KAMH \citep{sejdinovic2014kernel} and KHMC  \citep{strathmann2015gradient}. Finally, we apply our algorithm to Gaussian process classification on 
real-world datasets. 

We use RBF kernel $k(\bd{x}, \bd{x}')=\exp(-\|\bd{x}-\bd{x}'\|^2/h)$ for the updates of our proposed algorithms and the kernel approximation in \eqref{bt}; the bandwidth $h$ is taken to be $h {=} \mathrm{med^2}/(2\log(n+1))$ where $\mathrm{med}$ is the median of the current $n$ particles. When maximum mean discrepancy (MMD) \citep{gretton2012kernel} is applied to evaluate the sample quality, RBF kernel is used and the bandwidth is chosen based on the median distance of the exact samples so that all methods use the same bandwidth for a fair comparison. Adam optimizer \citep{kingma2014adam} is applied to our proposed algorithms for accelerating convergence.

\begin{figure*}[ht]
\centering
\begin{tabular}{ccccc}
\includegraphics[height=0.17\textwidth]{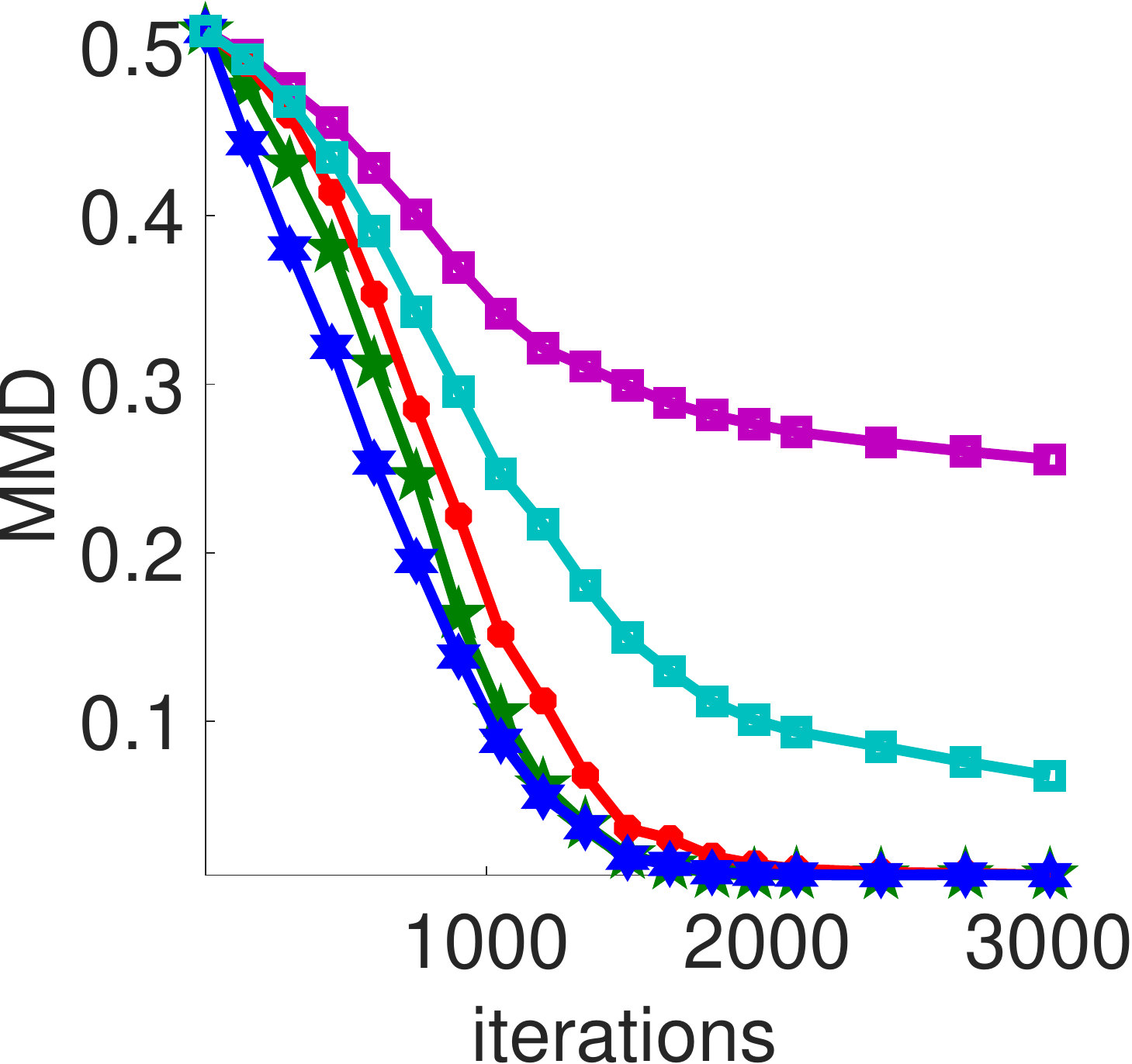}&
\includegraphics[height=0.17\textwidth]{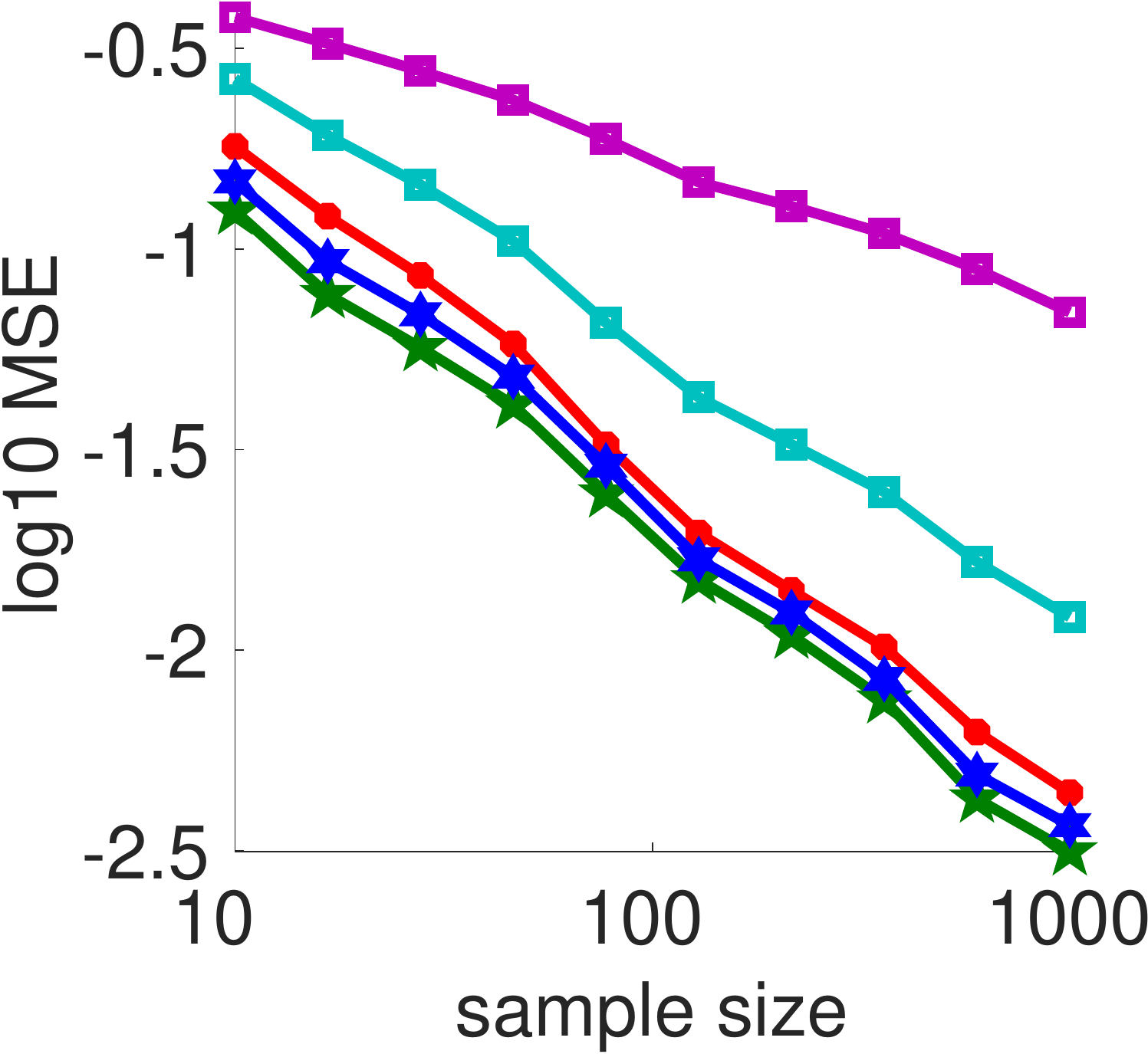}&
\includegraphics[height=0.17\textwidth]{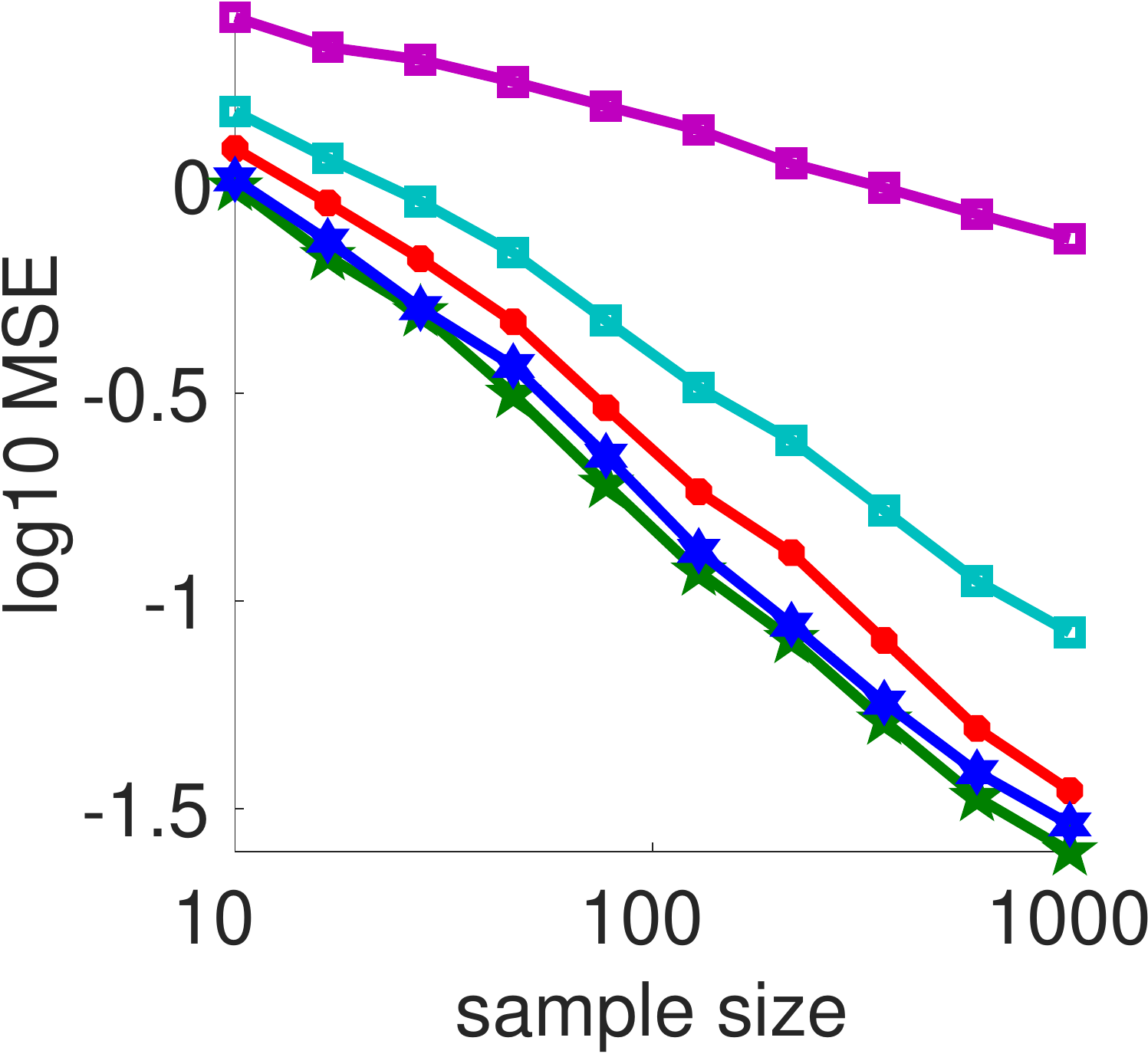} &
\includegraphics[height=0.17\textwidth]{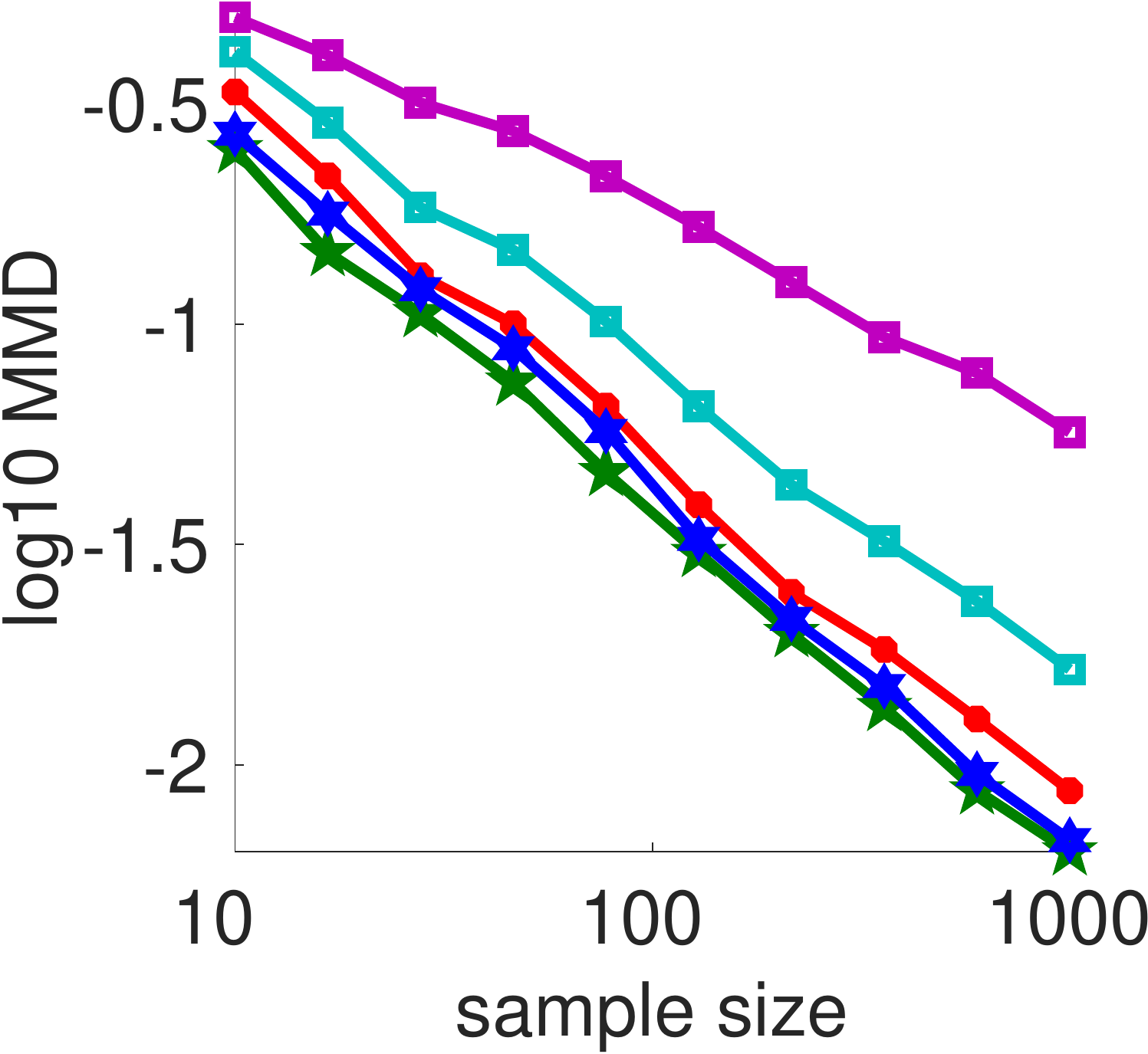}& 
\raisebox{2.0em}{\includegraphics[height=0.08\textwidth]{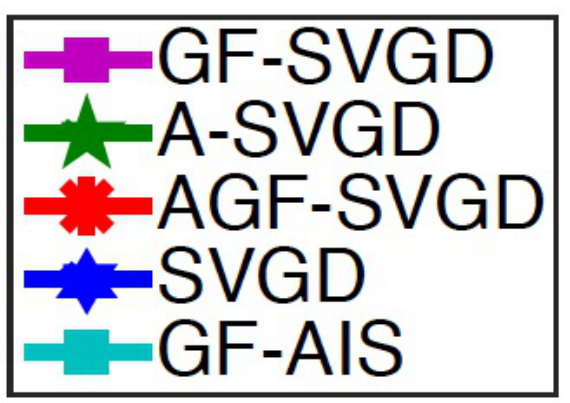}}\\
{\small (a) Convergence} & {\small (b) Mean} &  {\small (c) Variance} & {\small (d) MMD} \\
\end{tabular}
\caption{\small 
Results on GMM with 10 random mixture components and 25 dimensions. 
(a): the convergence of MMD with fixed sample size of $n=200$. 
(b)-(c): the MSE vs. sample size when estimating the mean and variance using the particles returned by different algorithms at convergence. 
(d): the MMD between the particles of different methods and the true distribution $p$. In (b, c, d), 3000 iterations are used.
For GF-AIS, the sample size $n$ represents the number of parallel chains, 
and the performance is evaluated using the weighted average of the particles at the final iteration with their importance weights given by AIS.}
\label{fig:gfgmm}
\end{figure*}

\subsection{Simple Gaussian Distributions}   
We test our basic GF-SVGD in Algorithm \ref{alg:alg1} 
on a simple 2D multivariate Gaussian distribution to develop insights on the optimal choice of $\rho$. 
We set a Gaussian target $p(\vx)=\mathcal{N}(\vx;\bd{\mu}, \sigma I)$ with fixed $\bd{\mu}=(0, 0)$ and $\sigma=2.0$, 
and an auxiliary distribution  $\rho(\vx)=\mathcal{N}(\vx;\bd{\mu}_\rho, \sigma_\rho I)$ where we vary the value of $\vv \mu_\rho$ and $\sigma_\rho$ in Figure~\ref{fig:gfgauss}. 
The performance is evaluated based on MMD between GF-SVGD particles and the exact samples from $p$ (Figure~\ref{fig:gfgauss}(a)), 
and the mean square error (MSE) of estimating $\vv\mu$ and $\sigma$ (Figure~\ref{fig:gfgauss}(b)-(c)).

Figure~\ref{fig:gfgauss} suggests a smaller difference in mean $\vv \mu_\rho$ and $\vv \mu$ generally gives better results,
but the equal variance $\sigma_\rho = \sigma$ does not achieve the best performance. 
Instead, it seems that $\sigma_\rho \approx 3 \sigma$ gives the best result in this particular case. 
This suggests by choosing $\rho$ to be a proper distribution that well covers the probability mass of $p$, it is possible to even outperform the gradient-based SVGD which uses $\rho=p$. 

Interestingly, even when we take $\sigma_\rho = \infty$, corresponding to the simple update in \eqref{invp} with $\rho=1$, 
the algorithm still performs reasonably well (although not optimally) in terms of MMD and mean estimation (Figure~\ref{fig:gfgauss}(a)-(b)). 
It does perform worse on the variance estimation (Figure~\ref{fig:gfgauss}(c)), 
and we observe that this seems to be because 
the repulsive force domains when $\sigma_\rho$ is large (e.g., when $\sigma_\rho=\infty$, only the repulsive term is left as shown in \eqref{invp}), and it causes the particles to be overly diverse, yielding an over-estimation of the variance. 
This is interesting because we have found that the standard SVGD with RBF kernel tends to underestimate the variance, 
and a hybrid of them may be developed to give a more calibrated variance estimation. 

In Figure \ref{fig:gfgauss}(d), we 
add additional comparisons with exact Monte Carlo (MC) which directly draws sample from $p$, and standard importance sampling (IS) with $\rho$ as proposal. 
We find that GF-SVGD provides much better results than the standard IS strategy with any $\rho$. 
In addition, GF-SVGD can even outperform the exact MC and the standard SVGD when auxiliary distribution $\rho$ is chosen properly (e.g., $\sigma_\rho\approx 3\sigma$). 


\subsection{Gaussian Mixture Models (GMM)}
We test GF-SVGD and AGF-SVGD on a 25-dimensional GMM with 10 randomly generated mixture components, 
$p(\vx)=\frac{1}{10}\sum_{i=1}^{10}\mathcal{N}(\vx;\bd{\mu}_i, I)$, 
with each element of $\vv \mu_i$ is drawn from $\mathrm{Uniform}([-1,1])$.  
The auxiliary distribution $\rho(\vx)$ is a multivariate Gaussian  $\rho(\vx) = \normal(\vx; \vv \mu_\rho, \sigma_\rho I)$, with fixed $\sigma_\rho = 4$ and each element of $\vv \mu_\rho$ drawn from $\mathrm{Uniform}([-1,1])$. 
For AGF-SVGD, we set
its initial distribution $p_0$ to equal the  
$\rho$ above in GF-SVGD. 

Figure~\ref{fig:gfgmm}(a) shows the convergence of MMD vs. the number of iterations 
of different algorithms with a particle size of $n = 200$, 
and Figure~\ref{fig:gfgmm}(b)-(d) shows the converged performance as the sample size $n$ varies. 
It is not surprising to see that that standard SVGD converges fastest since it uses the full gradient information of the target  $p$. 
A-SVGD converges slightly slower in the beginning but catches up later; this is because that it uses increasingly more gradient information from $p$. 
GF-SVGD performs significantly worse, which is expected because it does not leverage the gradient information. 
However, it is encouraging that annealed GF-SVGD, which also leverages no gradient information, performs much better than GF-SVGD, only slightly worse than the gradient-based SVGD and A-SVGD. 

For comparison, we also tested a gradient-free variant of annealed importance sampling (GF-AIS) \citep{neal2001annealed} with a transition probability constructed by Metropolis-adjusted Langevin dynamics, 
in which we use the same temperature scheme as our AGF-SVGD, and the same surrogate gradient $\nabla_{\vx} \log \rho_\ellt$ defined in \eqref{bt}. 
%
GF-AIS returns a set of particles with importance weights, so we use weighted averages when evaluating the MMD and the mean/variance estimation. 
This version of GF-AIS is highly comparable to our AGF-SVGD since both of them use the same annealing scheme and surrogate gradient.  %
However, Figure~\ref{fig:gfgmm} shows that AGF-SVGD still significantly outperforms GF-AIS. 
%

\begin{figure*}[ht]
\centering
\begin{tabular}{cccc}
\hspace{-.3cm} \includegraphics[height=0.18\textwidth]{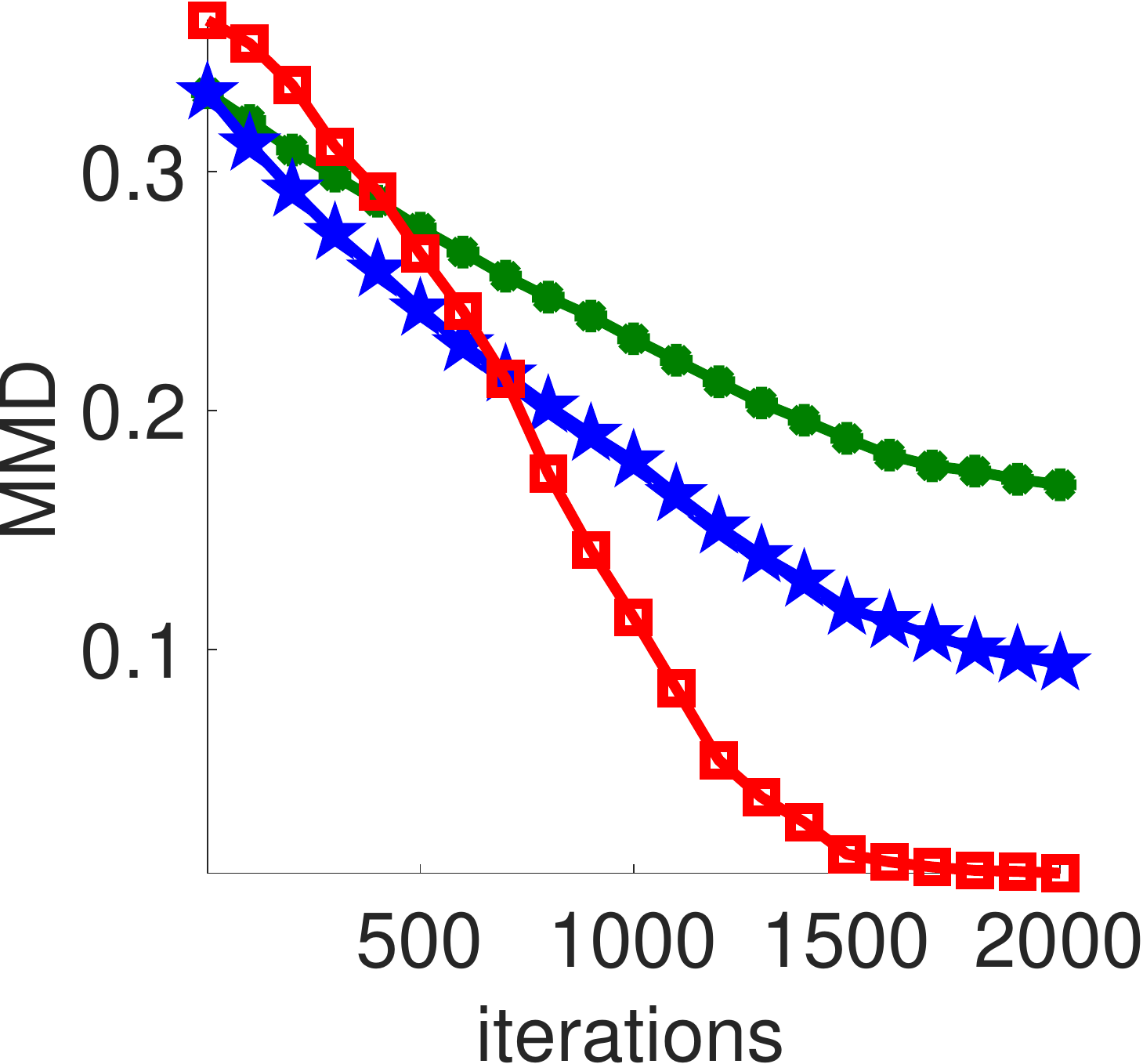}&
\includegraphics[height=0.18\textwidth]{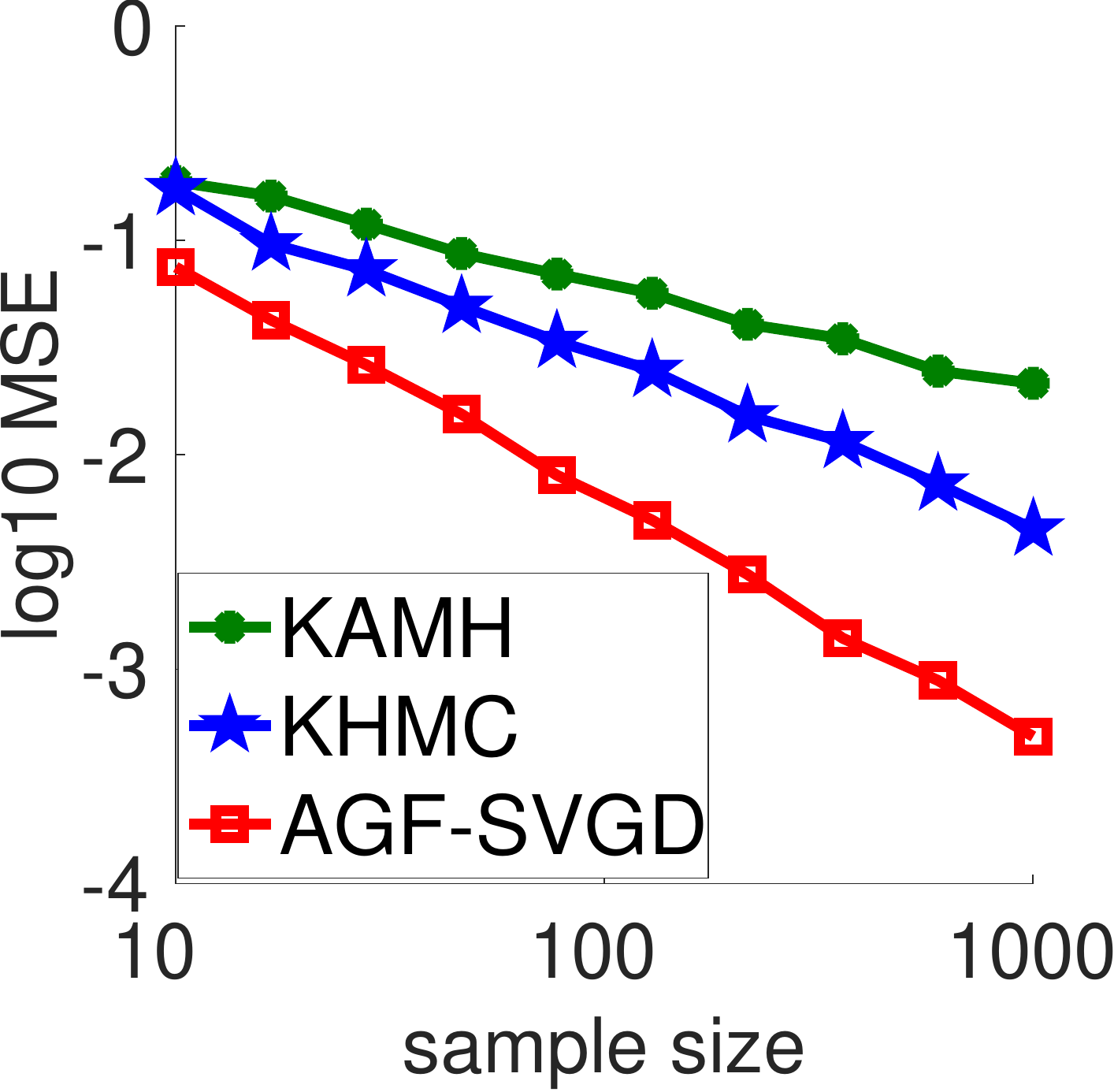} &
\includegraphics[height=0.18\textwidth]{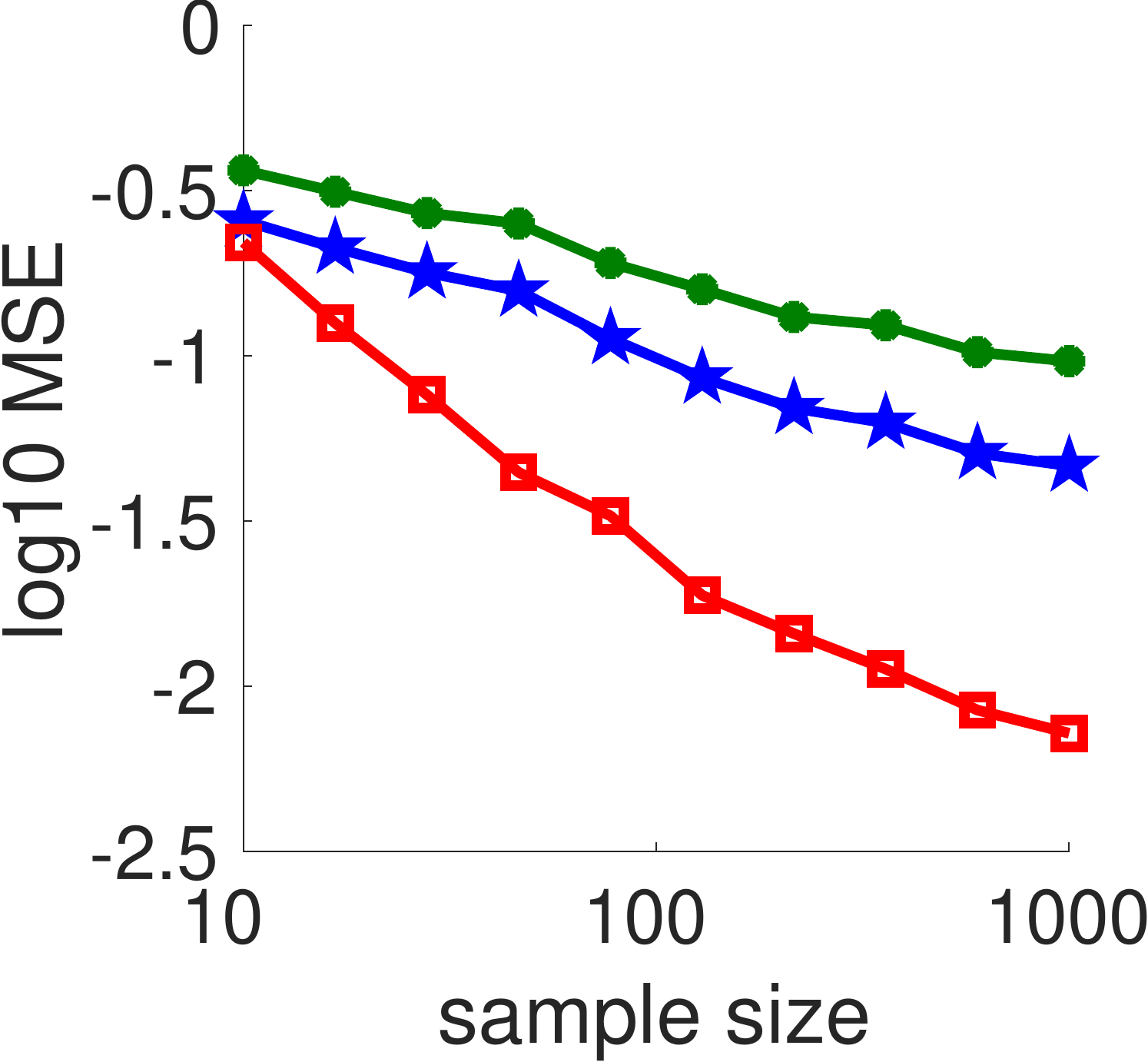} &
\includegraphics[height=0.18\textwidth]{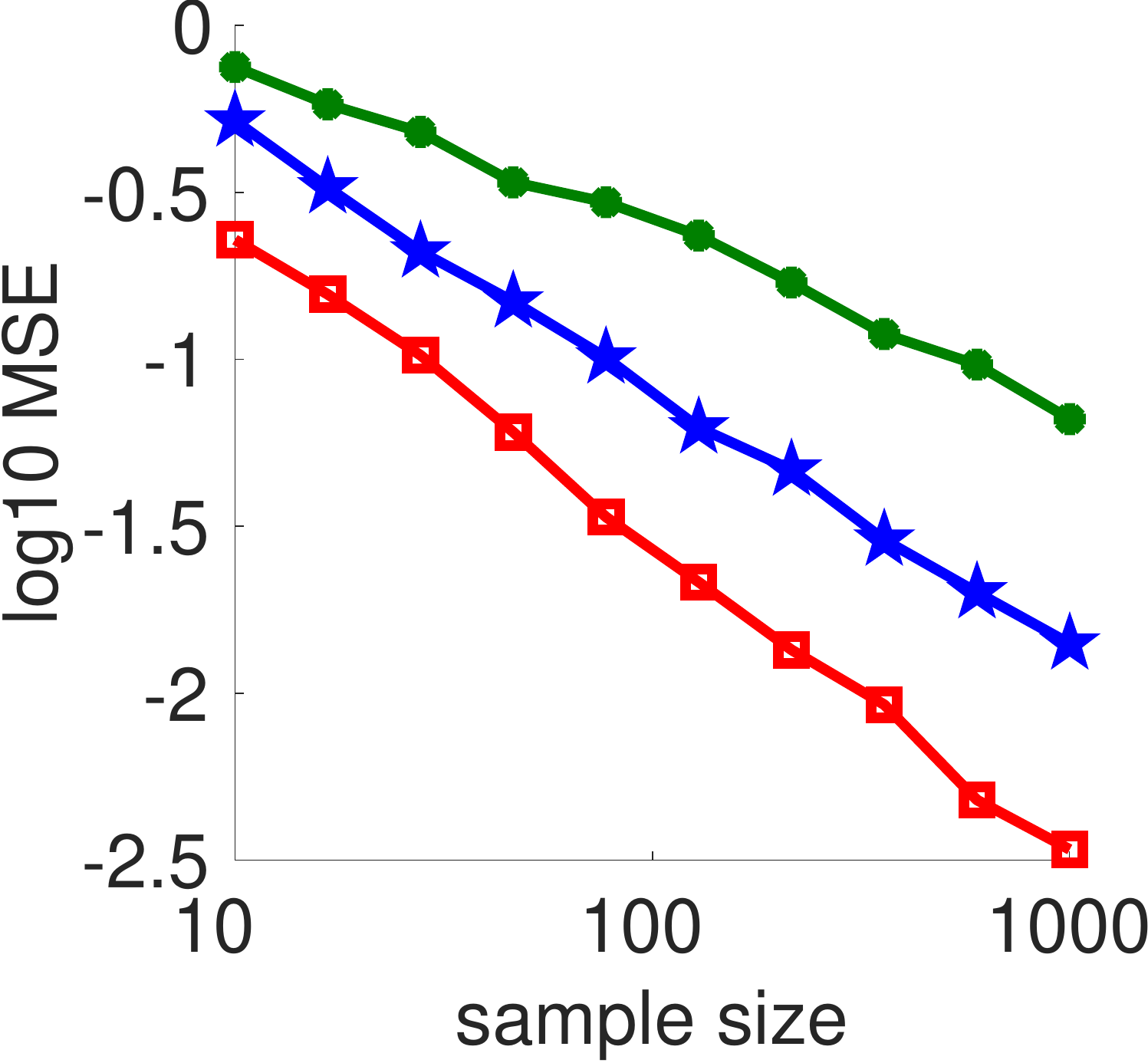}\\
 {\small (a) Convergence} & {\small (b) Mean} &  {\small (c) Variance} & {\small (d) MMD} \\
\end{tabular}
\caption{\small Gauss-Bernoulli RBM with $d=20$ and $d'=10$. (a): the convergence of MMD with $n=100$ for all the algorithms. The evaluations of MMD of KAMH and KHMC in (a) starts from the burn-in steps of the typical MH algorithm.
(b)-(c): the MSE vs. sample size when estimating the mean and variance using the particles returned by different algorithms at 2000 iterations.  
(d): the MMD between the particles of different algorithms and the true distribution $p$ at 2000 iterations.}
\label{Fig:rbm}
\end{figure*}

\subsection{Gauss-Bernoulli Restricted Boltzmann Machine}
We further compare AGF-SVGD with two recent baselines 
on Gauss-Bernoulli RBM, defined by 
\begin{equation}
p(\bd{x}) \propto \sum_{\vv h}\exp(\bd{x}^\top B\bd{h}+\vv c_1^\top\bd{x}+\vv c_2^\top\bd{h}-\frac{1}{2}\| \bd{x}\|_2^2),
\end{equation}
where $\vx\in \RR^d$ and $\vv h\in \{\pm1 \}^{d'}$ is a binary latent variable. 
By marginalizing the hidden variable $\vv h$, we can see that $p(\bd{x})$ is a special GMM with $2^{d'}$ components.  
In our experiments, we draw the parameters $\vv c_1$ and $\vv c_2$ from standard Gaussian and select each element of $B$  randomly from $\{\pm 0.5\}$ with equal probabilities. We set the dimension $d$ of $\vx$ to be 20 and the dimension $d'$ of $\vv h$ to be 10 so that $p(\vx)$ is a 20-dimensional GMM with $2^{10}$ components, for which it is still feasible to draw exact samples by brute-force for the purpose of evaluation. For AGF-SVGD, we set the initial distribution to be $p_0(\vx)=\mathcal{N}(\vx; \bd{\mu}, \sigma I)$, with  $\bd{\mu}$ drawn from $\mathrm{Uniform}([1, 2])$ and $\sigma =3.$

We compare our AGF-SVGD with two recent gradient-free methods: KAMH \citep{sejdinovic2014kernel} and KHMC \citep{strathmann2015gradient}. Both methods are advanced MCMC methods that adaptively improves the transition proposals based on kernel-based approximation from the history of Markov chains. 

For a fair comparison with SVGD, we run $n$ parallel chains of KAMH and KHMC and take the last samples of $n$ chains for estimation.  
%
In addition, 
we find that both KAMH and KHMC require a relatively long burn-in phase before 
the adaptive proposal becomes useful. In our experiments, we use 10,000 burn-in steps for both KAMH and KHMC, and exclude the computation time of burn-in when comparing the convergence speed with GF-SVGD in Figure~\ref{Fig:rbm}; this gives KAMH and KHMC much advantage for comparison, and the practical computation speed of KAMH and KHMC is much slower than our AGF-SVGD. 
%
From Figure~\ref{Fig:rbm} (a), we can see that our AGF-SVGD converges fastest to the target $p$, even when we exclude the 10,000 burn-in steps in KAMH and KHMC. 
Figure~\ref{Fig:rbm} (b, c, d) shows that our AGF-SVGD performs the best in terms of the accuracy of estimating the mean, variance and MMD. 

\subsection{Gaussian Process Classification}
We apply our AGF-SVGD to sample hyper-parameters from marginal posteriors of Gaussian process (GP) binary classification. 
Consider a classification of predicting binary label $y \in \{\pm1\}$ from feature $\vv z$. 
We assume $y$ is generated by a latent Gaussian process $f(\vv z)$, $p(y|\vv z) = 1/(1+\exp(-y f(\vv z )))$ and 
$f$ is drawn from a GP prior $f\sim GP(0, k_{f, \bd{\theta}})$, where $k_{f,\vv \theta}$ is the GP kernel indexed by a hyperparameter $\vv \theta$. In particular, we assume 
$k_{f,\vv{\theta}}(\vv z, \vv z') = \exp(-\frac12||(\vv z-\vv z')./\exp(\vv \theta)||^2)$, where $./$ denotes the element-wise division and $\vv\theta$ is a vector of the same size as $\vv z$. 
Given a dataset $Y = \{y_i\}$ and $Z = \{\vv z_i\}$, we are interested in drawing samples from the posterior distribution $p(\vv\theta|Z,Y)$. Note that the joint posterior of $(\bd{\theta}, f)$ is 
$$
p(\vv{\theta}, f | Z,Y) =  p(Y | f, Z) p(f|\bd{\theta}) p(\bd{\theta}). 
$$
Since it is intractable to exactly calculate the marginal posterior of $\vv\theta$, we approximate it by 
\begin{equation}
\label{estimate}
\hat{p}(\vv{\theta} |Z,Y):= p(\vv\theta)  \frac{1}{m}\sum_{i=1}^{m}
\frac{p(Y| f^i, Z) p(f^i |\bd{\theta}) }{q(f^i|\bd{\theta})}, 
\end{equation} 
where $\{f^{i}\}_{i=1}^m$ is drawn from a proposal distribution $q(f\mid\bd{\theta})$, which is constructed by an expectation propagation-based approximation of $p(f|\vv\theta,Z,Y)$ following %
\citet{filippone2014pseudo}.

\begin{figure}[ht]
\centering
\begin{tabular}{cc}
\includegraphics[height=0.18\textwidth]{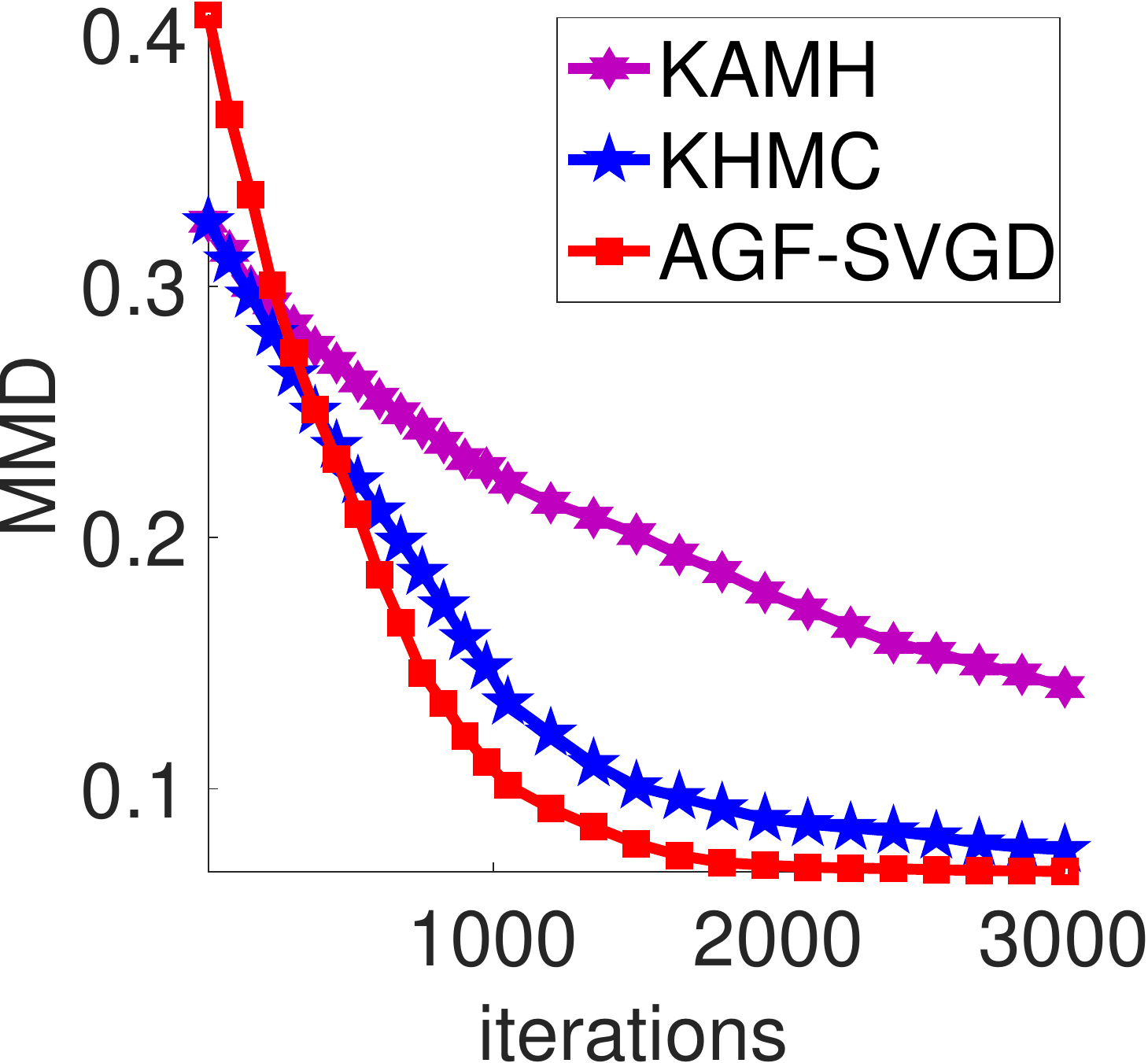} &
\includegraphics[height=0.18\textwidth]{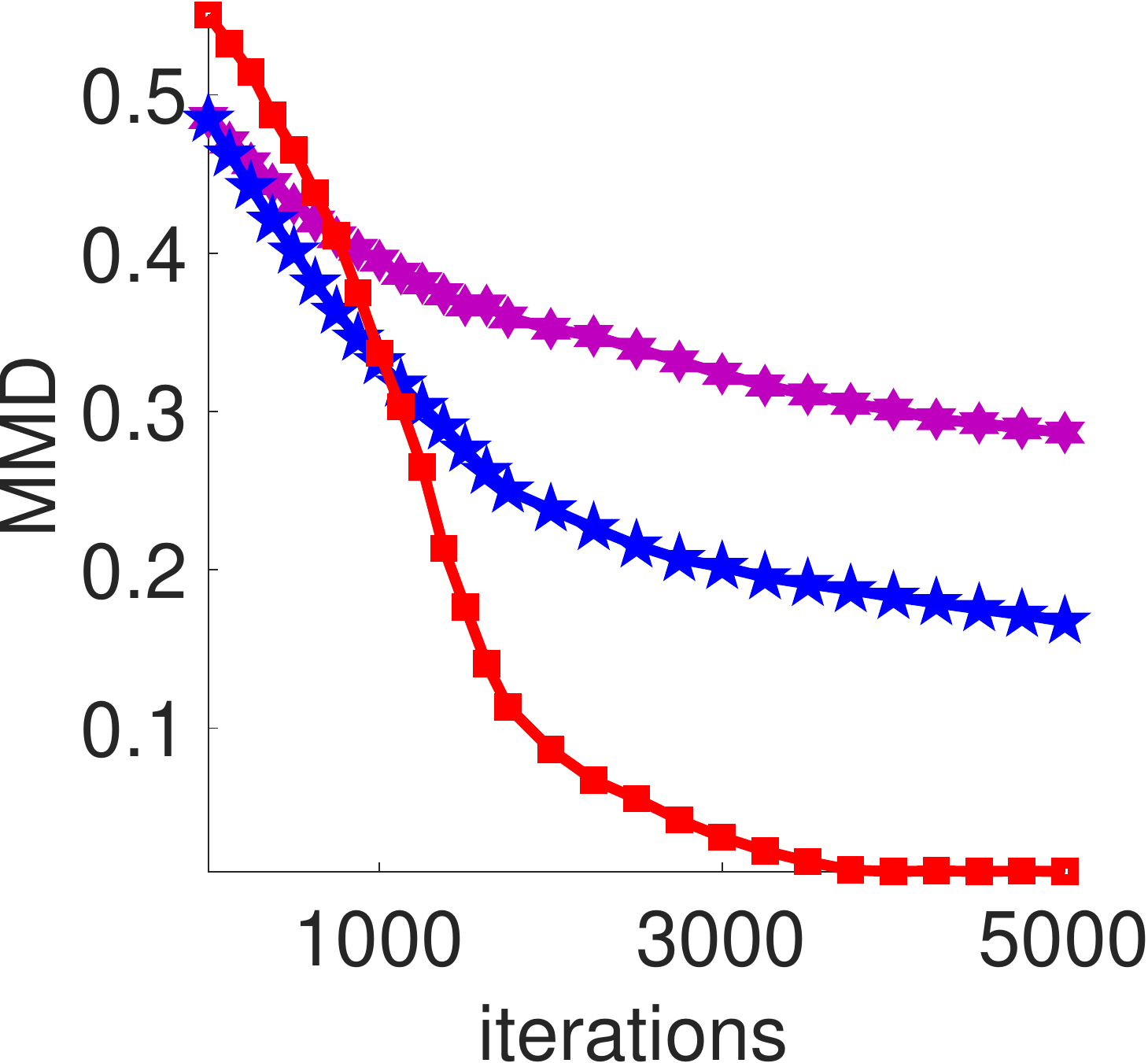} \\
{\small (a)  Glass dataset} &
{\small (b)  SUSY dataset} 
\end{tabular}
\caption{\small Sampling from the marginal posteriors on GP classification for Glass dataset (a) and SUSY dataset (b). We use a sample size of $n=200$ for all methods.}
\label{fig:real}
\end{figure}

We run multiple standard Metropolis-Hastings chains to obtain  ground truth samples from $p(\bd{\theta}\mid D)$, 
 following the procedures in section 5.1 of \citet{sejdinovic2014kernel} and Appendix D.3 of \citet{strathmann2015gradient}. 
We test the algorithms on Glass dataset and SUSY dataset in Figure~\ref{fig:real} from UCI repository \citep{asuncion2007uci} for which the dimension of $\bd{\theta}$ is $d = 9$ and $d=18$, respectively. 
We initialize our algorithm with draws from $p_0(\vx)=\mathcal{N}(\vx; \bd{\mu}, \sigma I)$ where $\sigma = 3$ and each element of $\bd{\mu}$ is drawn from $\mathrm{Uniform}([-1, 1])$. 
For KAMH and KHMC, 
we again run $n$ parallel chains and 
initialize them 
with an initial burn-in period of 6000 steps which {is \emph{not} taken into account in evaluation.}
Figure~\ref{fig:real} shows that AGF-SVGD again converges faster than KAMH and KHMC, even without the additional burn-in period.

\section{Conclusion}
We derive a gradient-free extension of Stein's identity and Stein discrepancy and propose a novel gradient-free sampling algorithm. 
%
Future direction includes theoretical investigation of optimal choice of the auxiliary proposal with which we may leverage the gradient of the target to further improve the sample efficiency over the standard SVGD.  We are also interested in exploring the application of the gradient-free KSD on the goodness-of-fit tests and black-box importance sampling. 

\section*{Acknowledgement}
This work is supported in part by NSF CRII 1565796.

\bibliographystyle{icml2018}
\bibliography{gradfree.bib}
\newpage
\onecolumn

\section*{Appendix A: Proof of Theorem \ref{pro:wphi}}
{\bf Proof:}, By definition, $w(\vx)=\rho(\vx)/p(\vx)$, $\nabla_{\vx}w(\vx)=w(\vx)\bd{s}_\rho(\vx)-w(\vx)\bd{s}_p(\vx)$,
\begin{align}
\steinpxtransp(w(\vx)\ff(\vx)) & = w(\vx) \bd{s}_p(\vx)^\top \ff(\vx) + \nabla_{\vx}^\top (w(\vx)\ff(\vx)) \notag\\
& = w(\vx) \bd{s}_p(\vx)^\top \ff(\vx) + \nabla_{\vx}w(\vx)^\top \ff(\vx) + w(\vx)\nabla_{\vx}^\top \ff(\vx) \notag \\
& = w(\vx) \bd{s}_\rho(\vx)^\top \ff(\vx) + w(\vx)\nabla_{\vx}^\top \ff(\vx) =w(\vx)\steinbxtransp \ff(\vx). \notag    
\end{align}
Therefore, we have 
\begin{align}
 \S_{\F, \prop}(q~||~p)  
 &  = \max_{\ff \in \F}\big\{ \E_{\vx\sim q} [\steinpxtransp \big( w(\vx)\ff(\vx)\big)]\big\} \label{app:sbf1} \\
 & = \max_{\ff \in w\F} \big \{\E_{\vx\sim q}[\steinpxtransp \ff(\vx)]]     \big \}  \label{app:sbf2}\\
 & = \S_{w\F}(q~||~p). \notag
\end{align}
\section*{Appendix B: Proof of Theorem \ref{theom}}
{\bf Proof:} When $\H$ is an RKHS with kernel $k(\vx,\vx')$, then $w\H$ is also an 
RKHS, with an ``importance weighted kernel'' 
\begin{align}\label{app:newkernel}
\tilde k(\vx,\vx') = w(\vx)w(\vx')k(\vx,\vx').
\end{align}
Following Lemma 3.2 in \citet{liu2016stein}, the optimal solution of the optimization problem  \eqref{app:sbf2} is,
\begin{align*}
w(\cdot){\ff}^*(\cdot) 
& = \E_{\vx\sim q}[\bd{s}_p(\vx) w(\vx)k(\vx, \cdot) w(\cdot) +\nabla_\vx (w(\vx)k(\vx, \cdot)w(\cdot))] \label{newvel}
\\
& = w(\cdot) \E_{\vx\sim q}[w(\vx) \steinbx k(\vx, \cdot)]. 
\end{align*}
This gives 
$$\ff^*(\cdot)= \E_{\vx\sim q}[w(\vx) \steinbx k(\vx, \cdot)].$$ 
Following 
Theorem 3.6 \citep{liu2016kernelized}, we can show that 
\begin{equation}
\label{app:newksd}
 \S_{\F, \prop}(q ~||~ p) = (\E_{\vx, \vx'\sim q}[\tilde{\kappa}_p (\vx,\vx')])^{\frac12},
\end{equation}
where 
$$
\tilde{\kappa}_p (\vx,\vx')
=(\stein_p')^\top(\stein_p\tilde k(\vx,\vx')).
$$
and $\stein_p$ and $\stein_p'$ denote the Stein operator applied on variable $\vx$ and $\vx'$, respectively. 
Applying Theorem~\ref{pro:wphi}, we have 
\begin{align*}
    \tilde{\kappa}_p (\vx,\vx')
& =(\stein_p')^\top\left (\stein_p (w(\vx)  w(\vx')k(\vx,\vx'))\right ) \\
& = (\stein_p')^\top(
 w(\vx)  \stein_\rho \left ( w(\vx')k(\vx,\vx'))\right ) \\
 & = (\stein_p')^\top(
 w(\vx') w(\vx)  \stein_\rho \left ( k(\vx,\vx'))\right ) \\
 &=  w(\vx') w(\vx)  (\stein_\rho')^\top(
 \stein_\rho \left ( k(\vx,\vx'))\right ) \\
 & =  w(\vx') w(\vx)  \kappa_\rho(\vx, \vx'),
\end{align*}
where we recall that $\kappa_\rho(\vx, \vx') = (\stein_\rho')^\top 
 \left (\stein_\rho k(\vx,\vx') \right)$. 
Therefore, $\D_{\F, \prop}(q, p)$ in \eqref{app:newksd} equals 
\begin{equation*}
\S_{\F, \prop}(q, p) =(\E_{\vx,\vx'\sim q}[w(\vx)\kappa_{\rho}(\vx, \vx')w(\vx')])^\frac12. 
\end{equation*}
This completes the proof. 
\hfill $\square$
\end{document}